# Earth Surface Immune System for Rapid Monitoring of Unknown Anomalies

Jingtao Li[a], Qian Zhu[b], Xinyu Wang[b], Deren Li[a], Liangpei Zhang[a], Yanfei Zhong[a*]

[a] State Key Laboratory of Information Engineering in Surveying, Mapping and Remote Sensing, Wuhan University, P. R. China.

[b] School of Remote Sensing and Information Engineering, Wuhan University, P. R. China.

*Corresponding author. Tel.: +86-27-6877996. E-mail address: zhongyanfei@whu.edu.cn.

**Abstract**: Earth surface anomalies, driven by escalating climate change, and expanding human activities, are increasing in both frequency and diversity, yet their limited historical data and unpredictability make them fundamentally different from conventional remote sensing targets. Existing methods address specific anomaly categories or stop at localization, leaving a gap between detection and actionable information. Here we present ESIA, an Earth Surface Immune System whose architecture is constrained by three principles from the biological immune system, refined over millions of years against equally diverse and uncertain threats. A non-specific innate immune stage treats anomalies as unobserved changes in time-series satellite imagery, generating binary localization maps at 14.51 $km^2/s$ without assuming any anomaly category, surpassing the strongest general baseline by 37% in $F_1$. A specific adaptive immune stage applies negative selection to filter text prompts and matches surviving prompts with localized image patches through a multi-modal foundation model, enabling open-vocabulary recognition of unknown anomaly attributes including category, affected area, and damage severity, with recognition $F_1$ exceeding 80%. A mutation mechanism tunes minimal embeddings at test time, adapting to each scene in 3.26s using a single reference image pair. We validate ESIA on a global-scale dataset covering 19,801.60 $km^2$ across six anomaly categories, comparing against 22 models, and further apply it to quantify degraded farmland in the Dnipro Delta following the Kakhovka Dam collapse and assess burn severity from 2025 Palisades Fire in Los Angeles. This unprecedented flexibility in handling unknown anomalies opens new avenues for real-time disaster response and environmental surveillance.

## 1. Introduction

Earth surface changes are pervasive on our dynamic planet, while only the anomaly changes have the vast power to reshape the Earth's terrestrial ecosystems (Noon et al., 2022; Qiu et al., 2025b). These changes, deviating from the historical balance state, alters the biophysical state substantially and poses a potential threat to the nature and human livelihood (Wei et al., 2023; Zhu et al., 2020). Human-directed anomaly changes, such as the non-agriculturalization, logging, and illegal urban expansion, together with the wild undirected ones, such as the wildfires, destructive storms, and drought, are key variables in the evolution of ecosystems for millennia. Exacerbated by the effects of global warming (Watson et al., 2023), there is an increasing frequency and intensity of anomaly changes on Earth surface (Domeisen et al., 2023; Yang et al., 2023), intertwined with the other problems of climate change (Xu et al., 2022). Natural disaster events caused approximately $224 billion in economic losses worldwide in 2025 and claimed over 17,000 lives, with extreme heat in the third hottest year on record resulting in at least 25,000 additional fatalities globally.

Monitoring the anomaly changes are largely different from the common remote sensing objects, where the unique properties of diversity, sample-scarcity, and the uncertainty make it a more challenging task (J. Li et al., 2024c; Zhu et al., 2020). Most studies focus on certain objects such as the damaged buildings after disaster (Shen et al., 2021; Zheng et al., 2021), non-agriculturalization activity (Sun et al., 2024), and the wild fire points (Wang et al., 2022), using carefully collected samples. In contrast, there is no such thing as "certain category" for anomaly changes, and any deviation from historical observation implying an anomaly change (J. Li et al., 2024b). This diversity makes it impossible to collect complete anomaly samples for supervised model training (Li et al., 2023; Xing et al., 2025). Additionally, we would never know the information about the next anomaly in advance, including the time, location, and the appearance, which is distinct from the post-disaster assessment task with known disaster type and location (Sarkar et al., 2023; United Nations Office for Disaster Risk Reduction, 2025).

These challenges make the monitoring of Earth surface anomalies a barren land in research communities. Almost all studies on anomaly changes assume the major impact on reduced vegetation (Yang et al., 2023; Zhu et al., 2020), and focus on the forest anomaly changes (Liu et al., 2023; Senf and Seidl, 2021; Ye et al., 2021). Classical vegetation indexes, like normalized difference vegetation index (NDVI), soil-adjusted vegetation index (SAVI), and green normalized difference vegetation index (GNDVI) are mostly used to model the healthy vegetation distribution in general anomaly detection methods (Castillo-Villamor et al., 2021). However, some of the anomalies, such as the mechanical changes, invasive species, and fire in non-vegetated areas are not related to the vegetation volume directly. Recent breakthroughs aim to get rid of this assumption and support more diverse anomalies. The first research line models the normal state using more spectral indexes such as normalized difference water index (NDWI), snow index (NDSI), and bare soil index for bare land (Mardian et al., 2021; Wei et al., 2023), beyond the vegetation-related indexes. The other research line transfers to model the historical observation by devising a statistical time-series models and fitting it with the satellite spectral observations (Qiu et al., 2025b; Shang et al., 2022; Zhu et al., 2020), treating the difference between the prediction and observed values as the anomaly degree. The fitted statistical model is kind of harmonic regression model to express the period rules without using any hand-crafted spectral index. Both lines represent the paradigms of prior knowledge-driven and data-driven, respectively. Despite the promotion in monitoring categories, the process is conducted with the original spectral signatures at pixel-level, making the results sensitive to the satellite image quality. Additionally, the harmonic regression process is computed per-pixel and per-band, leading to a heavy time cost.

Moreover, existing models are still stuck in the localization stage to output the anomalous locations, lacking the ability to recognize the important related attributes (Pickens et al., 2025; Qiu et al., 2025b), such as the anomaly category and affected land cover, which are necessary for first-response policy-making and management especially at the global-scale. The common paradigm, regardless of the model types: spectral index-based, statistical regression-based, or deep learning-based, models the distribution of normal objects first due to the rare anomaly samples (J. Li et al., 2024a, 2024c), and then adopts some distance metric to compute the continuous anomaly degree. The final threshed anomaly map is a binary map with 1 indicating anomaly and 0 indicting normal state. We can only know the anomaly location and need manual interpretation or re-training a recognition model to obtain more information. The reason why previous studies rarely touched the subsequent attribute recognition stage is that the unlimited diversity of anomalies is struggle for existing recognition models designed for certain categories (Ma et al., 2024; Wu et al., 2024), and this huge gap results existing anomaly monitoring models equal to localization models merely. If our ambition is set a complete framework or system for various Earth surface anomalies, not just a localization model, the expert knowledge from the remote sensing community alone is far from enough to guide the overall architecture design.

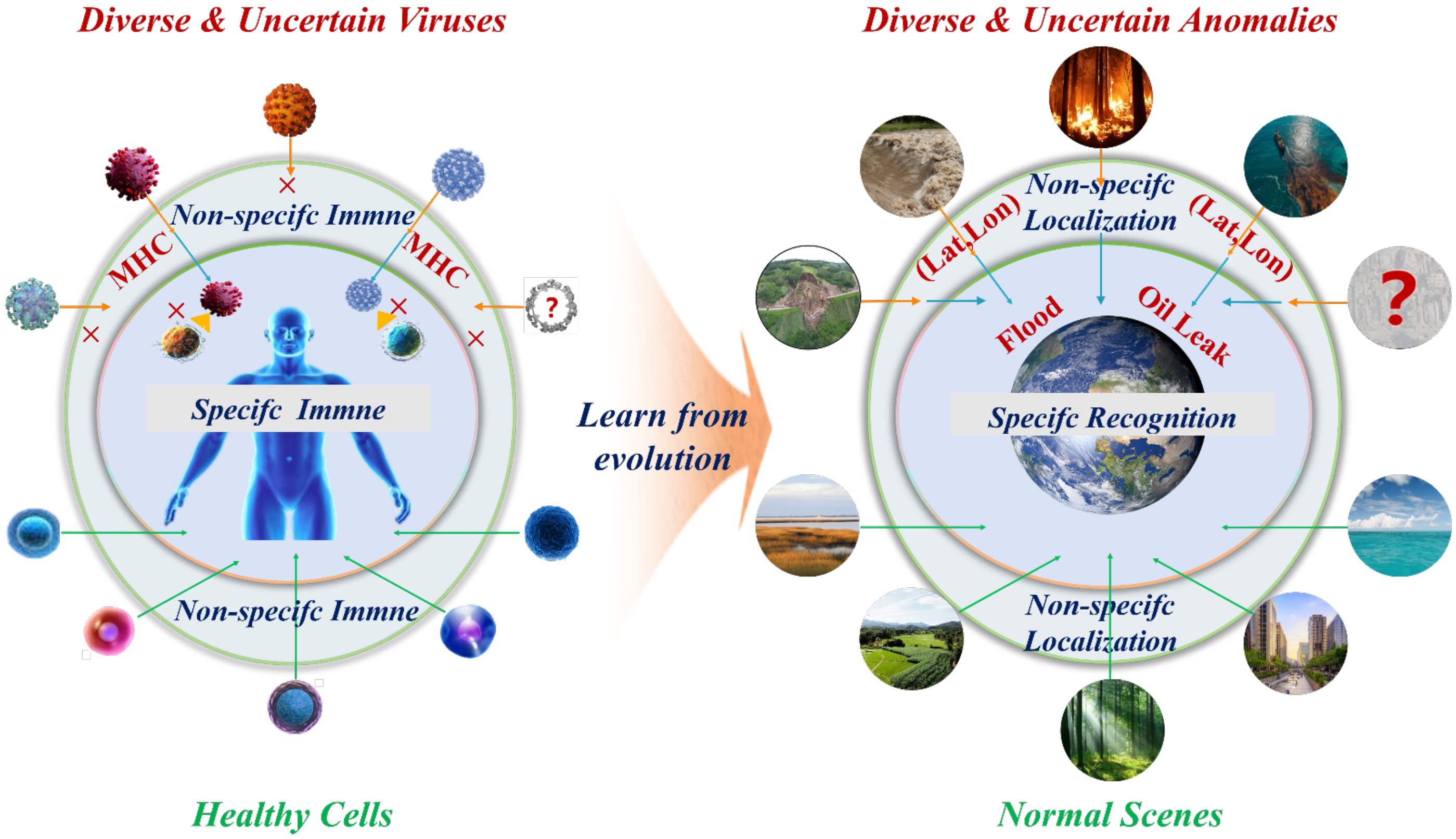

**Fig. 1.** We find that both viruses and Land surface anomalies exhibit similar properties of diversity and uncertainty, which inspires us to build the anomaly monitoring system via learning from biological immune system (BIS). BIS conducts non-specific immune first and upregulates the MHC level of viruses for latter specific immune, while proposed system designs corresponding non-specific anomaly localization first and records (latitude, longitude) for latter specific recognition stage, serving for diverse and uncertain Land anomalies.

Nature, however, has already solved a structurally analogous problem. The biological immune system (BIS), refined over 5 million years of evolution (Liston et al., 2021)., defends against viruses whose diversity and uncertainty mirror those of Earth surface anomalies, as in Fig. 1 (Shilts et al., 2022), inspiring us to propose a new path: *implementing the philosophy of BIS with modern techniques for Earth anomalies*. BIS, a multi-stage immune process (Domínguez-Andrés et al., 2022), includes non-specific innate immune as the first barrier and specific adaptive immune with complex mechanisms of negative selection, antibody-antigen matching, and the mutation. This staged architecture reflects an evolutionary principle that is absent from remote sensing practice, where end-to-end models pursue category-specific detection in a single pass and are limited to fixed categories. In contrast, for BIS, when the anomalies are unpredictable, non-specificity at the first stage can still work efficiently and the adaptive immune remains the recognition flexibility.

Our work presents a general and complete immune framework for Earth surface anomalies, named ESIA following the architecture of BIS. ESIA converts the monitoring target from viruses into time-series satellite images and implements the components in BIS correspondingly. The innate immune in ESIA is implemented as a zero-shot anomaly change detection task by treating anomalies as unobserved changes, generating non-specific binary localization map first with high processing speed 14.51 $km^2/s$, which acting the role of antigen-MHC complex in BIS (Nguyen and Youn, 2025). Then, the adaptive immune in ESIA locates each anomaly object with the binary map, and conducts specific recognition via the matching between image patches and text prompts (Guo et al., 2023), which is open-vocabulary (Zhu and Chen, 2024) and supports unlimited attribution recognition such as anomaly category and land use category. ESIA assigns the text prompts with highest matching degree as the recognition results. Like the T/B cells in BIS (Sender et al., 2023), the used text prompts in adaptive immune are generated by the built negative selection strategy to filter out the anomaly prompts with high-affinity binding to normal images. For the struggle samples, ESIA supports the real-time fast tuning to adapt to the image properties as the mutation in BIS (Harvey et al., 2021). ESIA enhances the generality of the innate immune utilizing spectral foundation model HyperFree (Li et al., 2025) and adaptive immune with multi-modality foundation model MS-CLIP (Jakubik et al., 2024). With all carefully designed components, ESIA can act as a real immune system for various anomalies, from non-specific localization to specific recognition. We construct a global-scale and time-series dataset covering 19801.60 $km^2$ to validate the effectiveness of ESIA, including six categories of dam break, burn area, fire point, marine debris, oil leak, and water bloom. Additionally, we apply ESIA to addressing two practical problems of global concern, the degraded farmland area in Dnipro Delta caused by the destroy of Kakhovka Dam in Russia-Ukraine war, and burn assessment at Palisades caused by the fire of Los Angeles in 2025, demonstrating the practical value of the BIS-guided architecture for first-response anomaly management.

## 2. **Study Area and Data**

ESIA is expected to be generalizable to various regions and anomaly types. We thus built a **g**lobal-scale **s**entinel-2 dataset covering 55 events for six types of Earth surface **a**nomalies (i.e., GSA dataset) as in Fig. 2(a), together with two local anomaly regions attracting global attention as in Fig. 2(e) and Fig. 2(f) to validate ESIA, where one corresponds to the affected farmland area in Dnipro Delta caused by the destroy of Kakhovka Dam in Russia-Ukraine war, and the other one corresponds to the burned area at Palisades caused by the fire in Santa Monica Mountains of Los Angeles on January 7, 2025. Time-series observation of sentinel-2 images are collected for each region to act as the practical monitoring data stream. We use the GSA dataset to conduct the quantitative evaluation on the innate immune and adaptive immune stages, module ablation and the prompt sensitivity analysis. Dnipro Delta is chosen to use ESIA answer the practical problem of “the impact of destroyed Kakhovka Dam on farmland at the Dnipro Delta, Ukraine", which is a problem of concern but still lack of quantitate answers. The collected diverse regions and anomaly types can show both the model robustness and promising application value of ESIA.

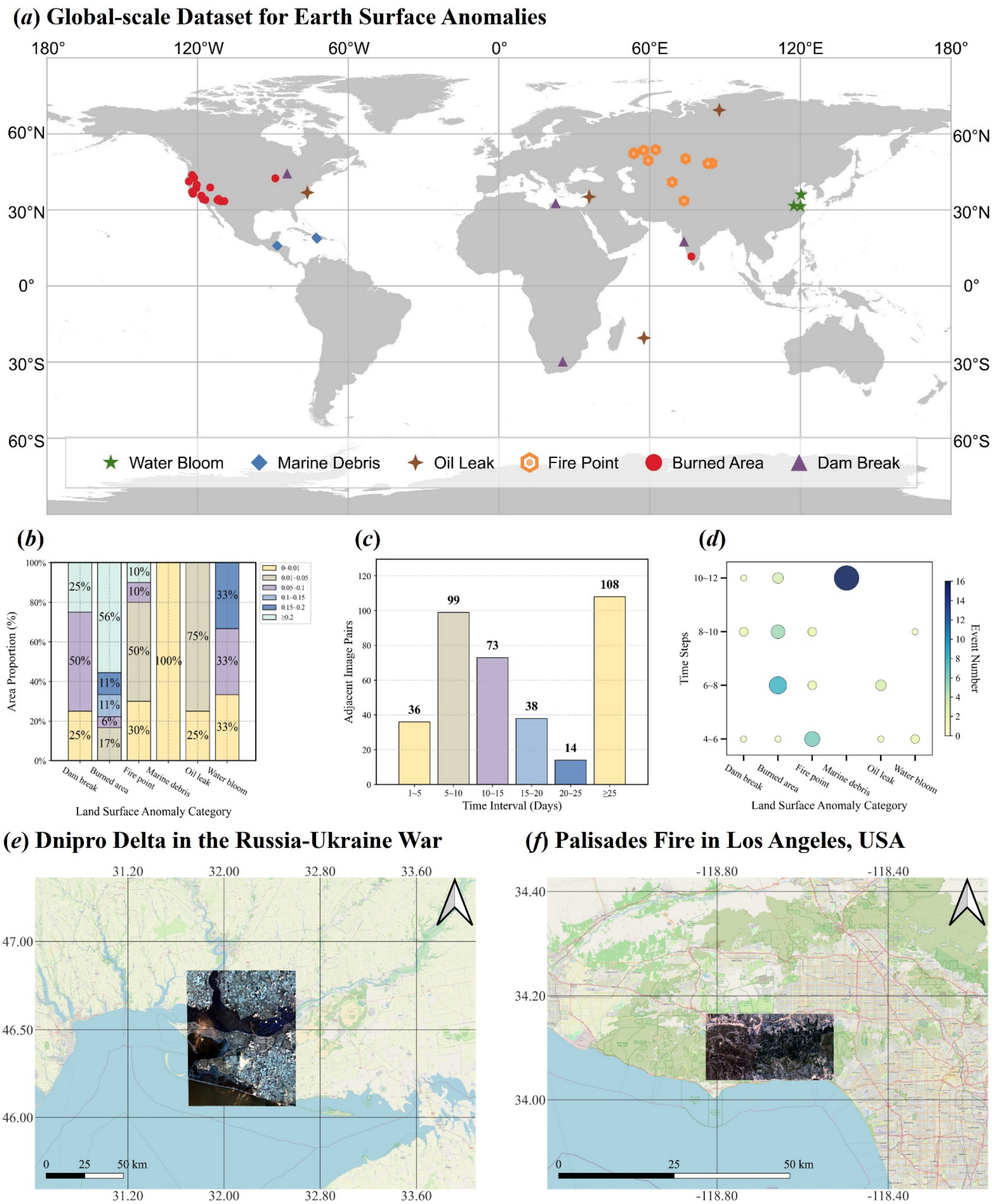

**Fig. 2.** We have built a global-scale sentinel-2 dataset covering 55 events for six anomaly categories (GSA) (a) and two local regions attracting global concern recently, where one is the affected Dnipro Delta caused by the destroy of Kakhovka Dam in Russia-Ukraine war (e), and the other one corresponds to the burned area at Palisades caused by the fire in Santa Monica Mountains of Los Angeles on January 7, 2025 (f). (b)-(d) are some statistics of GSA dataset.

## 2.1 Global-scale Dataset for Earth Surface Anomalies (GSA)

### 2.1.1 Dataset Collection and Annotation

We built GSA, a global-scale dataset covering 19801.60 $km^2$ for six anomaly categories as in Fig. 2(a) and Table 1, where the time-series observation for a total of 55 events was collected and annotated. Sentinel-2 images at level 2A are chosen as the data source because it achieves a satisfactory balance between the image resolution (i.e., 10~20 m/pixel), revisiting period (i.e., 5 days) and the spectral information (13 bands). The fusion of spatial-spectral information and the high revisiting frequency can help the continuous monitoring for various anomaly types even beyond the human visible. The bands of coastal aerosol, water vapour and SWIR-Cirrus are removed due to the mismatched low resolution of 60m/pixel, and there are 10 bands remained for each event. After obtaining the anomaly event information from the authoritative news report, we download one post-event image and unfixed number of pre-event images within the last six months for most events to form the series monitoring stream.

**Table 1**
The built GSA dataset covers 19801.60 $km^2$, with 55 events across 6 categories.

| Category | Description | Event number | Coverage area ($km^2$) |
|---|---|---|---|
| Dam break | Dam breaks can result from factors such as overtopping, foundation instability, or design flaws, leading to the unrestricted release of stored water or other materials. | 4 | 233.46 |
| Burned area | Land that has been affected by fire, typically resulting in charred vegetation and altered soil properties. | 18 | 17090.88 |
| Fire point | Fire point typically indicates the presence of ongoing combustion, such as a wildfire or controlled burn. | 10 | 398.72 |
| Marine debris | Marine debris refers to human-made waste—such as plastics, metals, or fishing gear—that ends up in oceans, seas, or other marine environments. | 16 | 104.86 |
| Oil leak | Oil leak refers to the unintended release of petroleum or crude oil into the environment, often into marine or coastal areas. | 4 | 147.21 |
| Water bloom | Water bloom is a rapid increase in the population of algae or cyanobacteria in freshwater or marine environments, often triggered by nutrient pollution. | 3 | 1826.46 |

We provide the pixel-level anomaly localization annotation and image-level anomaly category annotation for evaluating the both immune stages together. The annotation quality is guaranteed with the images from related news report and the cross-validation between different annotators. The anomaly events we collected are all influential and many news reports act as the ground annotation to provide the detailed location. Since some anomalies are struggle to distinguish the boundaries with natural RGB bands and we use some classical false color combinations and spectral indexes such as floating algae index (FAI) (Colkesen et al., 2024) to improve the quality. The anomaly category of marine debris comes from the research MARIDA (Kikaki et al., 2022), where we only choose the high-confidence annotation and complete the time-series images for the original single-temporal version. Three annotators are trained to operate with the same criterion and each event would be confirmed three times in turn. Once some inconsistency exists, the corresponding event need to be reannotated until all agreed with the pass. The annotated anomaly map is binary and involves various objects such as buildings, road, mountain, and the water, which is different from the most datasets with certain categories (Chen et al., 2025; Gupta et al., 2019). GSA would be made public along with the code of ESIA.

*2.1.2 Dataset Statistic*

We have reported some important statistics about the anomaly area ratio (Fig. 2(b)), the detailed revisiting periods (Fig. 2(c)) and number of time-series images (Fig. 2(d)). Due to the low-probability property of anomalies, most collected anomaly regions occupy the ratio lower than 10% in the monitoring images (Fig. 2(b)), making the detection task struggle. All the events in Marine debris category have ratio lower than 1% since the floating waste is very scattered with a few pixels, which shows a great difference between the category of burned area with 56% events more than 20%. Although the sentinel-2 satellite is designed for 5-days revisiting, the practical period has a certain difference when we calculated the adjacent time intervals of all the events in Fig. 2(c). 208 adjacent pairs have intervals less than 15 days and there are 108 image pairs with intervals larger than 25 days. Longer revisiting period implies an increased changing probability and detection difficulty. When forming the series observation stream, we limit the observation window of 6 months for most regions and the final number of time steps is reported in Fig. 2(d). 44 events have the observation data larger than 6 time steps and all the regions of the marine debris have time steps more than 10. The time-series observation data is consistent with the practical scenarios and offer a valuable reference information to recognize the anomalies in ESIA.

#### *2.1.3 Dataset Usage*

ESIA is significantly different from the existing models in terms of data usage, where it processes all the unseen images directly without training and builds the out-of-box immune procedure for usage and evaluation. This setting is more in line with emergent anomaly-response and imposes higher demands on model generalization. When carrying out the innate immune stage for each event, ESIA takes the post-event images of GSA as current observation and leverages all the historical steps to output pixel-level and non-specific anomaly localization maps for evaluation. For the adaptive immune stage, the situation is slightly different because it is defined as the image-level task and recognizes the anomaly category with single-temporal post-event images. Since post-event images of GSA are all positive samples, we build a sample pair with one pre-event image and one post-event image for each event to evaluate the adaptive immune. According to the property of human BIS, the binary location maps are used at the evaluation for non-specific innate immune stage and the anomaly class names for the specific adaptive immune stage.

### *2.2 Degraded Dnipro Delta in the Russia-Ukraine War*

During the ongoing Russia-Ukraine war, the Kakhovka Dam, located on the Dnipro River in southern

Ukraine, was destroyed on June 6, 2023. The resulting floods have caused serious damage to Ukrainian people's lives and property, animals and agriculture. 22,000 people were affected and 600 homes were flooded. The collapse of a dam has inundated tens of thousands of hectares of farmland in southern Ukraine and turned at least 500,000 hectares of underirrigated land into "deserts". The most severe long-term impact has been on the Dnipro Delta, a vital wetland ecosystem and important agricultural region. Due to the sudden drop in water level, the failure of the irrigation system and the soil salinization, the types of crops in Dnipro Delta have decreased significantly and a large amount of farmland has been left uncultivated.

To monitor these anomaly changes and give a quantitative evaluation for reference, we collect the related sentinel-2 data and apply ESIA to assess the area. Dnipro Delta mainly consists of the rice, corn and wheat, where the common harvest period is from July to August and the crop coverage can be clearly identified at that time. Therefore, we collect the series level 2-A sentinel-2 data from July to August in 2022, 2024 and 2025, forming the data stream with 6 time steps. The region size is 8597×10125 covering 8704.46 km$^2$. We use the data in 2022 as an uncorrupted reference and the data in 2024 and 2025 to confirm the lasting anomaly changes. Since there is on open-source ground information for the detailed degradation, our study on Dnipro Delta may only provide reference for researchers and policy-makers without the groundtruth evaluation.

### *2.3 Palisades Fire in Los Angeles, USA*

Los Angeles suffered its most severe fire in history at the beginning of 2025. Since January 7, 2025, a series of ongoing wildfires, exacerbated by extremely low humidity, drought and Santa Ana winds, caused damage to over 18,000 buildings and forced more than 200,000 people to evacuate. Pacific Palisades, source of the fire, was the most severely affected area, with the fire covering an area of 237.13 km$^2$.

Large-scale satellite observation has played an important role in monitoring the Palisades Fire and assessing the damage. We thus choose this practical and tragic case to validate the effectiveness of ESIA. We have collected five time steps from November 8, 2024 to January 2, 2025 as pre-event reference, and the image at January 12, 2025 as the post-event observation. All the images come from the level 2-A product of sentinel-2 platform and have the registered spatial size 3322×1418 covering 471.06 $km^2$. The groundtruth is made following the identical workflow in constructing the GSA dataset, and the bands of Near-Infrared (NIR) and Short-Wave Infrared (SWIR) are particularly utilized.

## 3. Methodology

### *3.1 Earth surface anomaly monitoring framework inspired by BIS*

The complex but efficient biological immune system (BIS) guides the proposed framework ESIA for Earth surface anomalies, and it is necessary to give a detailed introduction of BIS first (as in Fig. 3(a)). BIS is evolved into a multi-stage system and mainly includes the innate immune stage and adaptive immune stage (Liston et al., 2021). Given various cells in human body, the innate immune is the first barrier to prevent viruses relying on the primary effector cells such as phagocytes and natural killer (NK) cells, which can distinguish the healthy self-cells and viruses. The innate immune stage is fast and non-specificity (Shilts et al., 2022), but the defense intensity is relatively weak for all types of viruses. Therefore, the struggle viruses are always marked with antigen-MHC complex for the subsequent adaptive immune stage (Nguyen and Youn, 2025).

Adaptive immune targets at these marked viruses and exerts specific immune mainly with T-cells, B-cells, and the corresponding mutated ones (Deets and Vance, 2021). Negative selection plays an important role in generating massive and diverse T/B-cells while preventing the autoimmunity by deleting the T/B-cells with high-affinity binding to self-pMHC (peptide–Major Histocompatibility Complex) (Saurabh and Verma, 2023). When the virus comes in, the T/B-cells that can bind to viral antigens are activated and proliferated in large quantities. During proliferation, the genes encoding the antibody variable regions would undergo somatic

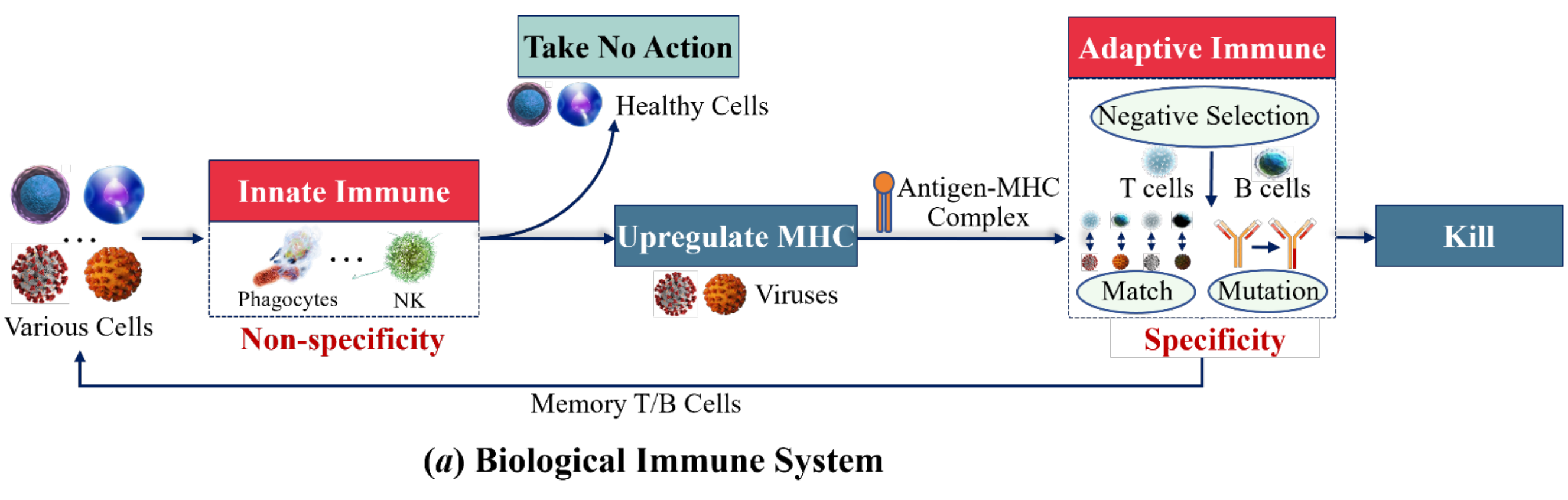

**(*a*) Biological Immune System**

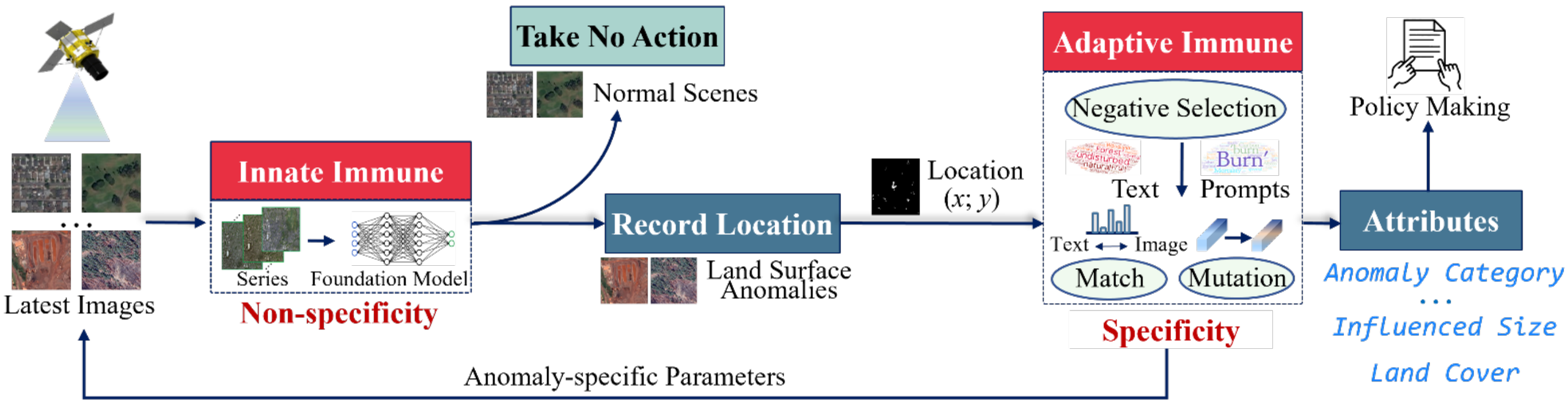

**(*b*) Earth Surface Immune System**

**Fig. 3.** The proposed Earth surface immune system corresponds to the main architecture of human immune system, including the innate immune and adaptive immune with components of negative selection, matching, and mutation. The difference lies at the inner instantiation for viruses and Earth surface anomalies, respectively.

hypermutation (SHM) (Gross, 2025), resulting in antibodies with higher affinity for viral antigens. Both T-cells and antibodies generated from B-cells are specific for the viral antigens through the matching process. The cooperation of both immune stages builds the powerful virus barrier and the generated memory T/B cells would enable a faster, stronger, and more specific immune response at next time.

We simulate the BIS system from its different key components to build ESIA as in Fig. 3 (b). The monitoring targets are changed from cells to remote sensing images covering diverse scenes. Innate immune, negative selection and adaptive immune are kept the identical term and relationship but with different inner procedures. To keep the non-specificity of the innate immune stage, ESIA models the normal distribution only with time-series images and the zero-shot models. Given latest unseen images, innate immune stage scores the anomaly degrees at pixel-level and highlights the locations of diverse Earth surface anomalies in the unsupervised and category-agnostic manner. The output binary format plays the similar role with antigen-MHC complex in BIS for latter specific adaptive immune. Negative selection in ESIA treats the ready-to-use prior knowledge (i.e., anomaly-related text prompts) as T/B-cells and can conduct specific recognition with matching process between the text prompts and localized image patches from innate immune stage. We enhance the recognition ability by devising a real-time prompt tuning strategy for each image patch, where the prompt description is allowed to be changed instantaneously rather than static according to the image properties like the mutation in BIS. Regardless of the different implementation techniques such as statistical or deep learning models, the built ESIA framework aims to support locating the diverse anomalies rapidly in innate immune stage and recognizing the specific attributes such as categories of anomaly and related land uses in adaptive immune stage.

### *3.2 Innate Immune Stage for Earth Surface Anomalies*

Given latest observation images, innate immune is the first step to give a rapid and non-specific response for diverse anomaly categories. Existing possible implementations mainly utilize the statistical models to identify the anomalous spectral signals (Castillo-Villamor et al., 2021; Senf and Seidl, 2021; Wei et al., 2023), which can generate non-specific outputs but with limited representation ability. In contrast, deep learning models always need the supervised training process with considerable training time and anomaly-specific response ability (Sarkar et al., 2023; Zheng et al., 2024). In our implementation, we tackle these fundamental problems and propose an efficient version for innate immune with time-series and frozen features from the foundation model as in Fig. 4.

#### *3.2.1 Anomalies are Unobserved Changes*

Traditional supervised training strategy makes the deep learning model being anomaly-specific and our implementation changes to identify the diverse anomalies from time-series observations but not the human annotation. The main idea behind is that anomalies must correspond to some changes that unobserved in the historical normal observations. On the other hand, once the changes were observed before and they are treated as uninteresting normal changes. Typical uninteresting changes are caused by period human activities, phenological changes and changes in imaging conditions (J. Li et al., 2024c). We thus treat the anomaly localization task in innate immune as a time-series comparison task, which is the same for all anomaly categories.

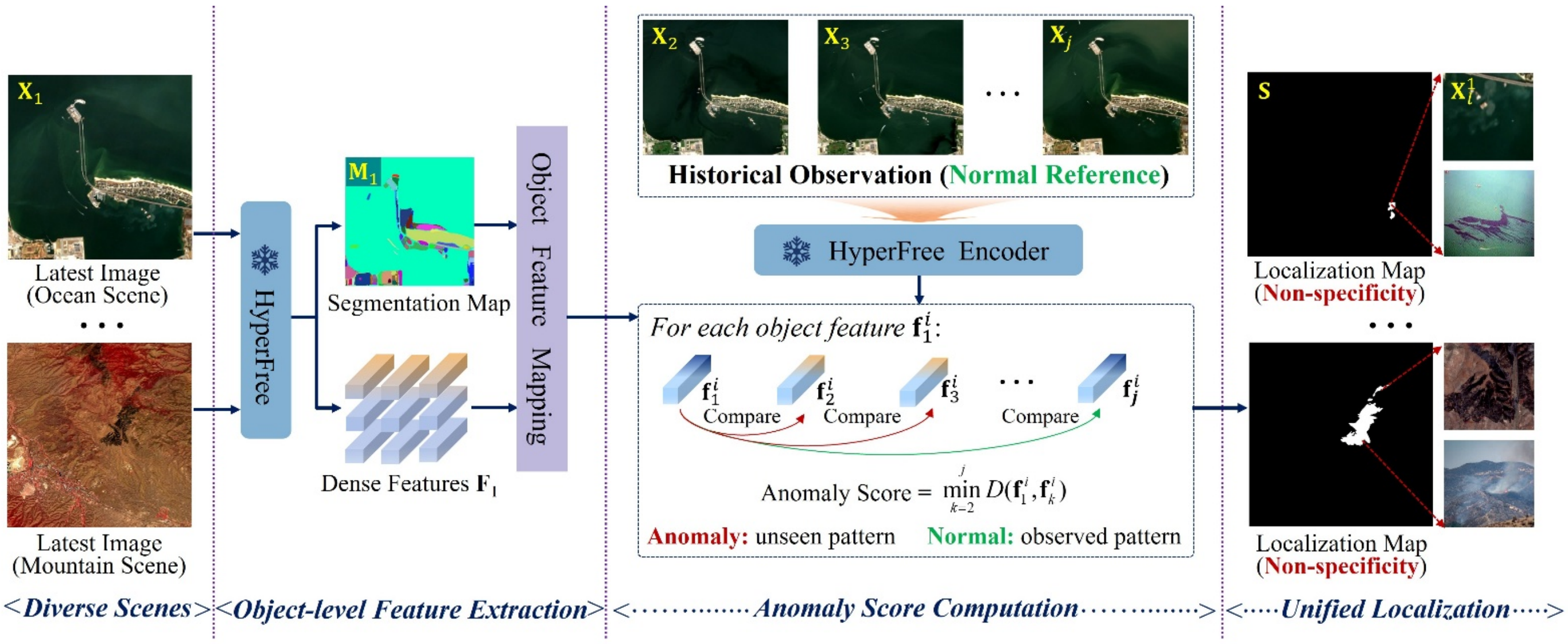

**Fig. 4.** The framework of innate immune stage in ESIA, which accepts the time-series observation images and outputs non-specific anomaly localization maps for diverse scenes and anomaly categories, treating the anomalies as unobserved changes.

### 3.2.2 *Time-series Comparison at Object-level*

We choose the feature space from spectral foundation model HyperFree (Li et al., 2025) to conduct the time-series comparison due to the unique properties of training-free and generalizable object segmentation. Monitoring the Earth surface anomalies always involve different satellites such as Landsat 8/9 and Sentinel 2. The full-spectrum embedding dictionary in HyperFree makes it support the dynamic embedding of these inconsistent wavelengths, which does not need the repeated and resource-intensive training processes. The training-free ability is also important for global-scale observation with great distribution shifts. Except for the strong backbone, HyperFree can output a dense object-level segmentation map at the same time. The map supports that we conduct the time-series comparison taking the ground objects as units rather than single pixels, suppressing the spectral heterogeneity problem. With HyperFree as the feature engine, users can utilize our innate immune system as simple as a statistical model.

Based on the above idea, we can formulate the innate immune stage with HyperFree for any scene and anomaly category. We have provided related visualization in Fig. 4. Set the latest monitoring image for some region as $\mathbf{X}_1 \in \mathrm{R}^{h \times w \times c}$, and the accumulated $j$ historical images as $\{\mathbf{X}_1, \mathbf{X}_2, ..., \mathbf{X}_j\}$. Our framework is compatible with different image height $h$, width $w$ and channels $c$ due to the adaptive ability of HyperFree. The revisiting period and time steps $j$ in historical images also allow for variability. We firstly process the $\mathbf{X}_1$ individually to get the backbone features $\mathbf{F}_1 \in \mathrm{R}^{64 \times 64 \times 256}$ and segmented masks $\mathbf{M}_1 \in \mathrm{R}^{h \times w \times n}$ including $n$ objects, where each object mask $\mathbf{m}_1^i \in \mathrm{R}^{h \times w}$ ( $0 \le i \le n$ ) is a binary map highlighting the object locations. If we further mapped the object location from $\mathbf{m}_1^i$ to $\mathbf{F}_1$, object-level feature vector $\mathbf{f}_1^i$ can be obtained where $\mathbf{f}_1^i = \text{mean}\, \mathbf{F}_1\, (\mathbf{m}_1^i == 1)$. After extracting image feature cubes $\{\mathbf{F}_1, \mathbf{F}_2, ..., \mathbf{F}_j\}$ for all historical images, we can generate object-level features $\{\mathbf{f}_1^i, \mathbf{f}_2^i, ..., \mathbf{f}_j^i\}$ at different time steps for the identical object in $\mathbf{X}_1$. The anomaly score for $\mathbf{m}_1^i$ can be finally computed as the minimum comparison score as in Eq. (1), where $D$ is the cosine distance metric.

$$S(\mathbf{m}_1^i) = \min_{k=2}^{j} D(\mathbf{f}_1^i, \mathbf{f}_k^i) \quad \text{where } \mathbf{f}_k^i = \text{mean}\, \mathbf{F}_k\, (\mathbf{m}_1^i == 1) \tag{1}$$

*3.2.3 Time-series Comparison at Pixel-level*

Eq. (1) outputs object-centric and non-specific anomaly maps for latter adaptive immune stage. It has shown robust detection ability in various categories in our experiments but fails in cases where only pixel anomalies exist, such as the marine debris in Sentinel-2 images (Kikaki et al., 2022), mostly occupying several pixels. To deal with such problems, we also design a pixel-level comparison method into our innate immune stage. We choose to model the historical observed spectra in certain location with classical multivariate gaussian distribution (MGD) since the deep learning models would introduce disruptive spatial features to weaken the tiny pixel signals. In such condition, the features $\{\mathbf{f}_1^i, \mathbf{f}_2^i, ..., \mathbf{f}_j^i\}$ in Eq. (1) is degraded into original spectral signals $\{\mathbf{X}_1^i, \mathbf{X}_2^i, ..., \mathbf{X}_j^i\}$ in image space. $D$ is instantiated as the Mahalanobis distance considering the varied scales in different wavelengths, where more historical images make a precise estimation of covariance matrix. The pixel-level anomaly map through time-series comparison can be formally expressed in Eq. (2).

$$S(\mathbf{X}_1^i) = D(\mathbf{X}_1^i, \mathrm{MGD}(\mathbf{X}_2^i, ..., \mathbf{X}_j^i)) \quad (2)$$

### *3.3 Adaptive Immune Stage for Earth Surface Anomalies*

BIS relies on the adaptive immune as the subsequent step of innate immune to exert the specific recognition and kill of various antigens. We follow this design in proposed ESIA framework, where we devise the corresponding adaptive immune to recognize the anomaly attributes based on the binary localization result from the innate immune stage. Achieving this target is difficult for traditional deep learning models since anomaly samples are always rare and insufficient to support the model learning (Zheng et al., 2024). Except for the sample problem, the trained models are mostly applicable to a fixed number of anomaly attributes (He et al., 2024; Qiao et al., 2023), forming a large gap with practical diverse and uncertain applications. ESIA solves the problem by simulating the matching process between antigens and antibodies (Liston et al., 2021), where we treat the open-vocabulary text prompts as antibodies and localized image patches from innate immune stage as antigens to conduct the matching process. Negative selection and prompt mutation strategies are correspondingly developed to make the adaptive immune in ESIA being complete and robust for various anomalies, like a real BIS as in Fig. 5.

### 3.3.1 Treating Diverse Text Prompts as Antibodies

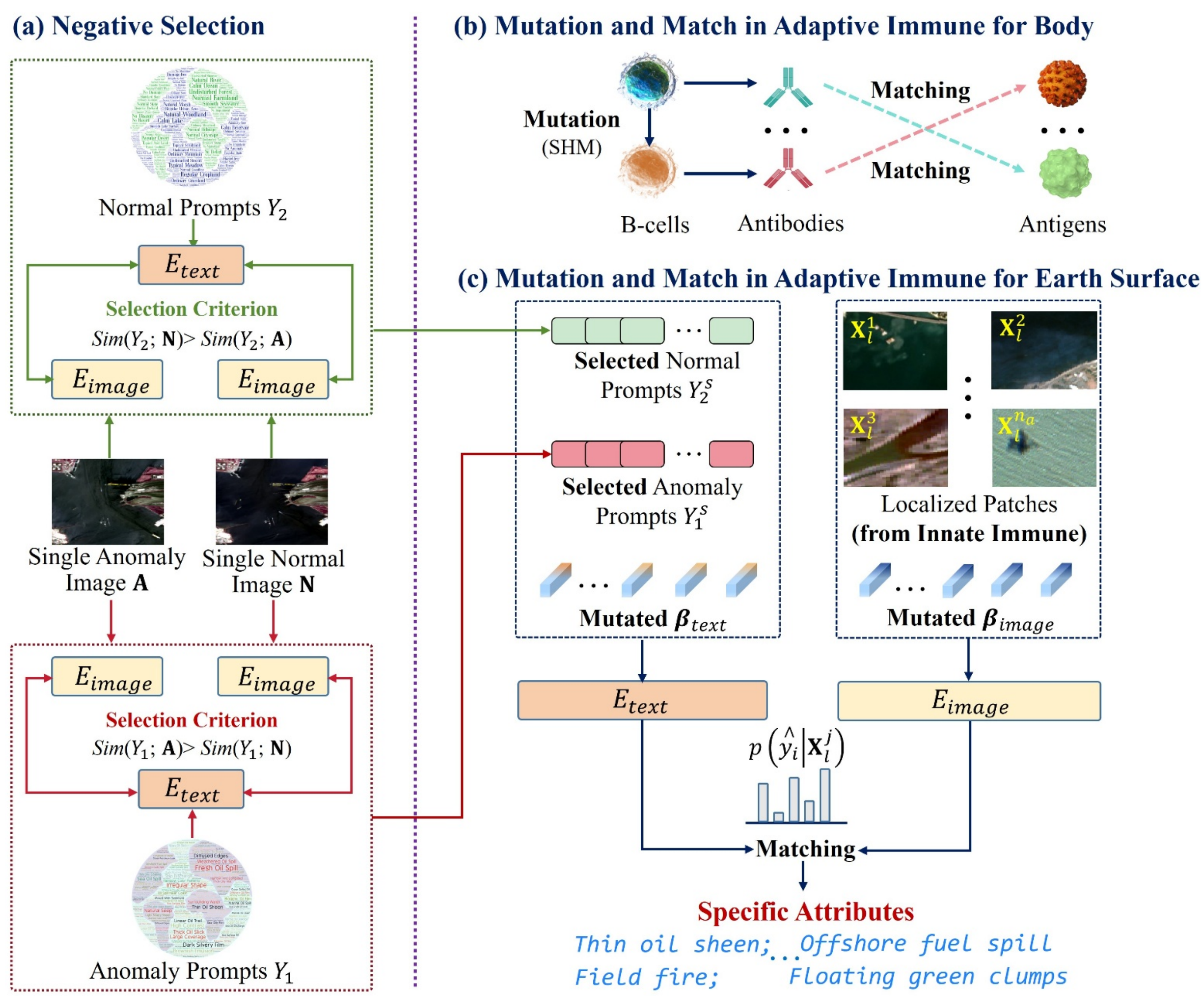

**Fig. 5.** The framework of adaptive immune stage in ESIA, which processes the localized local anomaly patches from innate immune stage, and recognizes the specific anomaly attributes such as category and size in text formats. Treating diverse text prompts as antibodies, we implement the important components of negative selection, mutation and match, corresponding to the design guidance from biological immune system.

To recognize diverse even unseen anomaly attributes like antibodies, we shift the recognition model design from traditional unimodal paradigm to the text-image multimodal matching paradigm (Radford et al., 2021) inspired by the matching in BIS. In practice, users wonder the anomaly attributes in many aspects such as the category, size and related land use category. Each attribute $Y$ can be expressed as a combination of all the possible text phrases $Y = \{y_1, y_2, ..., y_k\}$, where the $k$ phrases are always mutually exclusive for the one-winner matching. For examples, the text phrases would be {"Normal scene", "Fire Point", …, "Oil Leak"} for the category attribute and be {“Buildings”, “Ice”, …, “Grass”} for the land use attribute. For the image branch, previous anomaly localization map $S$ has filtered out most normal regions with a pre-set threshold following AnomalyCD (J. Li et al., 2024c) and the remaining anomaly objects can be specifically located with surrounding context. Assume there are $n_a$ potential anomaly objects in $S$ and the corresponding localized patches are denoted as $\{\mathbf{X}_l^1, \mathbf{X}_l^2, ..., \mathbf{X}_l^{n_a}\}$. Given the desired attribute $Y = \{y_1, y_2, ..., y_k\}$ and localized anomaly patches $\{\mathbf{X}_l^1, \mathbf{X}_l^2, ..., \mathbf{X}_l^{n_a}\}$ to be processed, we have two pretrained encoders $\{E_{text}, E_{image}\}$ from MS-CLIP (Jakubik et al., 2024) to extract the text and vision features respectively. Take some localized patch $\mathbf{X}_l^j$ as example, we encode the text features as $\{\boldsymbol{t}_1, \boldsymbol{t}_2, ..., \boldsymbol{t}_k\}$ for attribute $Y$ and image features as $\boldsymbol{v}_j$ for patch $\mathbf{X}_l^j$. Using the cosine metric $sim$ to measure the matching degree, we can then compute the attribute recognition probability as Eq. (3).

$$p(\hat{y}_i \mid \mathbf{X}_l^j) = \frac{\exp(sim(E_{image}(y_i) \bullet E_{text}(\mathbf{X}_l^j)))}{\sum_{i=1}^{k} \exp(sim(E_{image}(y_i) \bullet E_{text}(\mathbf{X}_l^j)))} = \frac{\exp(sim(\boldsymbol{t}_i \bullet \boldsymbol{v}_j))}{\sum_{i=1}^{k} \exp(sim(\boldsymbol{t}_i \bullet \boldsymbol{v}_j))} \quad (3)$$

Each desired attribute can generate its own matching vector of $\{p(\hat{y}_1 \mid \mathbf{X}_l^j), p(\hat{y}_1 \mid \mathbf{X}_l^j), ..., p(\hat{y}_k \mid \mathbf{X}_l^j)\}$ and obtain the most suitable description with argmax operation. The matching process of Eq. (3) is identical for all the attributes and anomaly patches keeping the $\{E_{text}, E_{image}\}$ frozen, and the category number is flexible due to the open-vocabulary property of pre-trained encoders (Yao et al., 2023).

### *3.3.2 Negative Selection for Preventing Autoimmunity*

Anomaly category may be the most concerned attribute in practice and there are numerous text phrases to act as the anomaly prompts $Y_1 = \{y_1, y_2, ..., y_j\}$ and $Y_2 = \{y_{j+1}, y_{j+2}, ..., y_{j+k}\}$. For example, we can describe a normal scene with prompts as “Normal land”, “No damage”, “Common status” or “Anomaly-free” and an oil leak event with prompts as “Sea oil spill”, “Marine surface oil” or “Dark floating slick”. These prompts can be generated in large quantities with large language models (LLM) (Naveed et al., 2025) and expert priors, like the massive T/B cells in human body. However, given a localized anomaly patch $\mathbf{X}_l^j$, not all the anomaly prompts have higher matching degree with it than normal prompts due to the bias problem of pre-trained $\{E_{text}, E_{image}\}$. Negative selection for category recognition prompts $Y = \{Y_1, Y_2\}$ is thus designed to prevent the autoimmunity problem in the prompt-image matching as Eq. (3). We would take the specific oil leak recognition problem to describe the process, which is identical for other anomaly categories and attributes.

Consistent with the BIS, our negative selection is conducted before the matching with real anomaly patches $\{\mathbf{X}_l^1, \mathbf{X}_l^2, ..., \mathbf{X}_l^{n_a}\}$ as a kind of preparation. BIS uses self-cells to filter out the T/B cells with high affinity and we introduce a paired of registered images $\{\mathbf{A}, \mathbf{N}\}$ to filter the prompts $Y = \{Y_1, Y_2\}$. $\mathbf{A}$ and $\mathbf{N}$ correspond to the same location except that one has anomalies (e.g., oil leak) and one does not, which makes the difference of prompt-image matching degree is only related to the anomaly patterns. For each anomaly prompt $y_i$ in $Y_1$, it can only be selected when it has higher affinity with $\mathbf{A}$ than $\mathbf{N}$ as Eq. (4), and similar criterion for normal prompts $Y_2$ as Eq. (5).

$$Criterion\ 1:\ sim(E_{text}(y_i), E_{image}(\mathbf{A})) > sim(E_{text}(y_i), E_{image}(\mathbf{N})) \quad y_i \in Y_1 \tag{4}$$

$$Criterion\ 2:\ sim(E_{text}(y_i), E_{image}(\mathbf{A})) < sim(E_{text}(y_i), E_{image}(\mathbf{N})) \quad y_i \in Y_2 \tag{5}$$

We extend the selection in BIS with only normal samples (i.e., self-cells) to the selection with both $\mathbf{A}$ and $\mathbf{N}$. Since there is not an absolute indicator as self-MHC (Duke-Cohan et al., 2023) to show the matching success, we have to use the $\mathbf{A}$ to form the ranking order. With frozen model parameters, our selection process is efficient to generate qualified prompts for the matching with real anomaly recognition process.

*3.3.3 Real-time Prompt Mutation for Strong Recognition*

Mutation is a masterpiece of BIS evolution to resist new and powerful viruses such as the SHM procedure (Gross, 2025). The selected prompts from Eq. (4) and Eq. (5) are denoted as $Y_1^s$ and $Y_2^s$ respectively ($Y_1^s \subset Y_1$ and $Y_2^s \subset Y_2$). SHM changes the immunoglobulin variable region with high frequency and ESIA allows $Y_1^s$ and $Y_2^s$ to change similarly. For each embedding $Em(y_i)$ of text prompt $y_i$, we concatenated an identical changeable embedding $\boldsymbol{\beta}_{text}$ to it as $\{\boldsymbol{\beta}_{text}; Em(y_i)\}$ before feeding into the $E_{text}$, where $Em$ represents the frozen embedding layer. $y_i$ is kept frozen while $\boldsymbol{\beta}_{text}$ is designed to adapt the prompt information for more accurate matching ability. To make the vision branch have similar ability to be adaptive, we further insert a changeable embedding $\boldsymbol{\beta}_{image}$ for each localized anomaly patch $\mathbf{X}_l^j$ and generate $\{\boldsymbol{\beta}_{image}; Em(\mathbf{X}_l^j)\}$ to vision encoder $E_{image}$. With the frozen $\{E_{text}, E_{image}\}$ and mutable $\{\boldsymbol{\beta}_{text}, \boldsymbol{\beta}_{image}\}$, the recognition probability $p(\hat{y}_i \mid \mathbf{X}_l^j)$ can be re-expressed from Eq. (3) to Eq. (6).

$$p(\hat{y}_i \mid \mathbf{X}_l^j) = \frac{\exp(sim(E_{image}(y_i, \boldsymbol{\beta}_{image}) \bullet E_{text}(\mathbf{X}_l^j, \boldsymbol{\beta}_{text})))}{\sum_{i=1}^{k} \exp(sim(E_{image}(y_i, \boldsymbol{\beta}_{image}) \bullet E_{text}(\mathbf{X}_l^j, \boldsymbol{\beta}_{text})))} \tag{6}$$

$\{\boldsymbol{\beta}_{text}, \boldsymbol{\beta}_{image}\}$ allows for the adjustment according to the properties of attribute and localized images. However, learning optimal $\{\boldsymbol{\beta}_{text}^*, \boldsymbol{\beta}_{image}^*\}$ is non-trivial since it needs to be fast enough and only single paired images $\{\mathbf{A}, \mathbf{N}\}$ (used in negative selection) and the $\mathbf{X}_l^j$ to be processed are available. It would not be an effective immune system for Earth surface if unlimited samples and time are permitted. The optimal target is given in Eq. (7).

$$\boldsymbol{\beta}_{text}^*, \boldsymbol{\beta}_{image}^* = \underset{\boldsymbol{\beta}_{text}, \boldsymbol{\beta}_{image}}{\arg\min} L(\mathbf{A}, \mathbf{N}, \mathbf{X}_l^j, Y_1^s, Y_2^s, \boldsymbol{\beta}_{image}, \boldsymbol{\beta}_{text}) \tag{7}$$

We aim to utilize all the possible supervision up to achieve the fast converge, and $L$ is designed as two parts with $L_{sup}$ and $L_{uns}$. $L_{sup}$ represents the supervised loss with $\{\mathbf{A}, \mathbf{N}\}$, which acts as a correct reference to guide the mutable embedding learning. In the context of oil leak recognition problem, we mark an oil leak as 1 if it occurs and 0 if it doesn't. Since lots of prompts are provided for $Y_1^s$ and $Y_2^s$, we take the maximum in each prompt set as the final prediction (i.e., $p(1 \mid \mathbf{A}) = \max(p(y_i \mid \mathbf{A}))$ for $y_i \in Y_1^s$ and $p(0 \mid \mathbf{A}) = \max(p(y_i \mid \mathbf{A}))$ for $y_i \in Y_2^s$). Together with the binary cross entropy loss (BCE) loss $L_{bce}$, we also employed pixel-level explainable losses $L_{exp}$ to ensure that the model located the true anomalous regions.

We denote the obtained heatmap of $\mathbf{A}$ as $\phi(\mathbf{A}) \in \mathrm{R}^{h \times w}$, where $\phi$ represents the procedure proposed by Li et al. (2024). $L_{exp}$ expects a large difference exist for $\mathbf{A}$ between the mean response value of anomaly regions (i.e., $Mean_1(\phi(\mathbf{A}))$) and normal regions (i.e., $Mean_0(\phi(\mathbf{A}))$), and a small difference for $\mathbf{N}$. The overall $L_{sup}$ is defined in Eq. (8).

$$L_{sup} = L_{bce} + L_{exp} = -\log(\max_{y_i \in Y_1^s}(p(y_i \mid \mathbf{A}))) - \log(\max_{y_i \in Y_2^s}(p(y_i \mid \mathbf{N}))) \\ + \mid Mean_1(\phi(\mathbf{N})) - Mean_0(\phi(\mathbf{N})) \mid - \mid Mean_1(\phi(\mathbf{A})) - Mean_0(\phi(\mathbf{A})) \mid \tag{8}$$

Unsupervised $L_{uns}$ is designed related to the properties of image $\mathbf{X}_l^j$ to be detected, which guides the embedding learning on the prediction consistency across different views. Specifically, a total of $U$ views of $\mathbf{X}_l^j$ are generated with a composition of data augmentation techniques such as rotation and contrast transformation. $\{\boldsymbol{\beta}_{text}, \boldsymbol{\beta}_{image}\}$ is required to mutate and made the consistent prediction despite the distribution varies. We use the average prediction entropy to measure the consistency and the $L_{uns}$ is defined in Eq. (9). The overall learning loss is the weighted sum of $L_{sup}$ and $L_{uns}$ in Eq. (10) with $\lambda$ to control the balance.

$$L_{uns} = -\sum_{y_i \in \{Y_1^s, Y_2^s\}} \frac{1}{U} \sum_{i=1}^{U} p(y_i \mid Aug_i(\mathbf{X}_l^j)) \log(\frac{1}{U} \sum_{i=1}^{U} p(y_i \mid Aug_i(\mathbf{X}_l^j))) \tag{9}$$

$$L(\mathbf{A}, \mathbf{N}, \mathbf{X}_l^j, Y_1^s, Y_2^s, \boldsymbol{\beta}_{text}, \boldsymbol{\beta}_{image}) = L_{sup}(\mathbf{A}, \mathbf{N}, Y_1^s, Y_2^s, \boldsymbol{\beta}_{text}, \boldsymbol{\beta}_{image}) + \lambda L_{uns}(\mathbf{X}_l^j, Y_1^s, Y_2^s, \boldsymbol{\beta}_{text}, \boldsymbol{\beta}_{image}) \tag{10}$$

Note that the optimization process for the mutated $\{\boldsymbol{\beta}_{text}, \boldsymbol{\beta}_{image}\}$ is different from the traditional tuning process, and our optimization is conducted just within several steps (typical 10~15 steps less than 6 seconds). This design is not only to make the mutation process effective but also maintain the open-world recognition ability of pre-trained models. Besides, the light mutated parameters of $\{\boldsymbol{\beta}_{text}, \boldsymbol{\beta}_{image}\}$ can be stored without pressure and loaded directly for next-time recognition, like the memory T/B-cells in BIS.

## 4. Experiments and analysis

### *4.1 Experimental settings*

**Use of Dataset**: ESIA keeps consistent with the environment of BIS, where it processes all the satellite images directly without a separate training stage as most models. Therefore, GSA and two local regions in Ukraine and Los, Angeles do not provide the subset of training. The mutation in adaptive immune stage tunes the mutable embeddings $\{\beta_{text}, \beta_{image}\}$ at the test time.

**Comparison Methods**: ESIA involves the interaction of a range of different models including anomaly localization, change detection, and anomaly recognition, which are all tested and compared to provide a comprehensive evaluation. For the non-specific innate immune stage, there are 14 anomaly-specific anomaly localization models, 4 bi-temporal unsupervised change detection models, and 4 general anomaly localization models that are used for comparison. Anomaly-specific anomaly localization models use hand-crafted spectral indexes to detect the anomalies, and we have reported the detailed formula in Table X. Active fire detection with sentinel-2 images (AFD-$S_2$) is a special one with multi designed criteria for the category of fire point and the remaining models belong to the spectral indexes (Hu et al., 2021); The chosen unsupervised change detection models belong to the category of statistical or zero-shot detectors, and we do not include the unsupervised detectors that are trained image-by-image from the fair and practical views. Change vector analysis (CVA) (Bovolo and Bruzzone, 2006), structure consistency-based graph (SCG) (Sun et al., 2021), and multivariate alteration detection (MAD) (Nielsen, 2007) are detectors designed for the bi-temporal multispectral images and have been widely used in communities. SAM-Bi is recently proposed for high-resolution Earth surface anomalies and we compare with it due to its unique property of zero-shot inferring (J. Li et al., 2024c), where the RGB bands are extracted from multi-spectral images to feed the network; In the past few decades, the general anomaly localization models are mostly developed with time-series spectral images (e.g., Landsat and Sentinel 2) and statistical methods (Qiu et al., 2025b; Zhu et al., 2020). Treating the historical images as normal, Continuous monitoring of Land disturbance (COLD) models the normal state as a multivariate normal distribution with original spectral signals (Zhu et al., 2020), while Earth observation-based anomaly detection (EOAD) (Castillo-Villamor et al., 2021) and prior knowledge-based anomaly detection (PKAD) (Wei et al., 2023) models with extracted spectral indexes such as NDVI, EVI, and SAVI. Based on the prior distributions, some distance metrics like Z-score and Mahalanobis distance, are always used to compute the anomaly degree for the latest image.

**Table. 1.** Used spectral indices for specialized comparison, where the grey indices represent that they are computed as the difference between bi-temporal images.

| **MNDWI** (Xu, 2006) | (B3-B11)/(B3+B11) | **dNDVI** (Jiang et al., 2006) | (B8-B4)/(B8+B4) | **TAI** (Xia et al., 2018) | (B11-B8)/(B11+B8) |
|---|---|---|---|---|---|
| **MBWI** (Wang et al., 2018) | 2×B3- B4- B8- B11-B12 | **dNBR** (Lutes et al., 2006) | (B8-B12)/(B8+B12) | **NDFI** (Liu et al., 2025) | (B12-B4)/(B12+B4) |
| **AWEI** (Jiang et al., 2014) | 4×(B3-B11)-(0.25×B8+02.75×B12) | **NDOI** (Brekke and Solberg, 2005) | (B5-B4)/(B5+B4) | **AFD-S²** (Hu et al., 2021) | Criteria (B4, B11, B12) |
| **FDI** (Alì et al., 2024) | B8-(B4+( B11 – B4)×0.1797) | **SABI** (Alawadi, 2010) | (B8-B4)-(B2+B3) | **FAI** (Hu, 2009) | B4+( B11 – B4)×0.2116 |
| **PI** (Themistocleous et al., 2020) | B8/ B4 | **RdNBR** (Miller and Thode, 2007) | $(dNBR - dNBR_{min})/(dNBR_{max} - dNBR_{min})$ | | |

When building the benchmark of the adaptive immune stage, the available comparison models are relatively scare since there are few studies achieving the recognition for a variety of anomaly categories. Therefore, different recognition models are separately compared for each anomaly category, and we convert the 14 anomaly-specific anomaly localization models in Table X to the recognition task with a threshold about the number of anomaly pixels. Marine debris is a special case where the extremely debris proportion makes the threshold difficult to set. Since ESIA uses the test-time adaptation technique in adaptive immune, we also tested some advanced adaptation methods on the identical base model MS-CLIP (Jakubik et al., 2024), including Context optimization (CoOp) (Zhou et al., 2022), test-time prompt tuning (TPT) (Shu et al., 2022) and in-context prompt learning (InCPL) (Yin et al., 2025). The prompt images $\{\mathbf{A}, \mathbf{N}\}$ and hyperparameters are kept the same for all the adaptations.

**Evaluation Criteria**: We access both the localization performance in innate immune and recognition in adaptive immune with the classical criteria (i.e., recall, precision, and $F_1$), which supports both the pixel-level and image-level. Recall indicates the proportion of damaged pixels that have been detected, precision shows the proportion of correctly detected damaged pixels, and $F_1$ provides a comprehensive evaluation. Denoting the number of true positive, false positive and false negative as TP, FP, and FN, respectively. Recall (R), precision (P) and $F_1$ metrics can be computed as below.

$$\mathrm{R} = \frac{\mathrm{TP}}{\mathrm{TP}+\mathrm{FN}} \tag{10}$$

$$\mathrm{P} = \frac{\mathrm{TP}}{\mathrm{TP}+\mathrm{FP}} \tag{11}$$

$$\mathrm{F}_1 = 2\frac{\mathrm{P}\cdot\mathrm{R}}{\mathrm{P}+\mathrm{R}} \tag{12}$$

**Implementation Details**: In the innate immune stage, we use a fixed sliding window with size 512 to process large-scale images, and 2% linear stretching is applied before feeding into the HyperFree to generate high-quality object masks. Both IoU and stability thresholds are set 0.4. In the adaptive immune stage, we generate the text phrases for each category with the Qwen3-max language model (Yang et al., 2025) for the negative selection and recognition. At the test time, the mutation for each image contains 35 iterations, with 30 supervised steps and 5 unsupervised steps. $\lambda$ is set as 0.4 to make the loss terms work together. Adam optimizer with learning rate 5e-3 and default betas are used to update the mutation parameters.

## *4.2 Global-scale Anomaly Monitoring Results*

### *4.2.1 Benchmark Results at Innate Immune Stage*

ESIA outputs non-specific anomaly localization maps for various categories, and we compare it with anomaly-specific models, general bi-temporal change detection models, and general time-series anomaly localization models. We reported the quantitative results together in Table 2. Various input types, including original reflectance, vegetation Indexes, and pre-trained features of deep learning models, are evaluated under the zero-shot setting for the fair and comprehensive comparison. The anomaly-specific models mostly use the hand-crafted spectral priors, and we implemented the classical ones for each category, such as the RdNBR for burned area and the AFD-$S^2$ for the fire point (Hu et al., 2021). We found most of these anomaly-specific models showed promising results and surpassed many general change detection results. For example, FDI (Alì et al., 2024) achieves the highest $F_1$ score 19.44 on the category of marine debris, which is not an easy thing due to the extreme low proportion (lower than 0.01) of debris. RdNBR (Miller and Thode, 2007) obtained the suboptimal results on the category of burned area with $F_1$ 62.13, close to the optimal result 66.97 by PKAD (Wei et al., 2023). Despite this, due to the high uncertainty of anomaly category and location, anomaly-specific results have inherent limitation in monitoring unseen regions. Bi-temporal change detection models, treating changed regions as anomaly regions, performed worse than the other two types of models due to the lack of ability to distinguish the anomaly changes from normal changes. When introducing time-series satellite data to model the normal state, an obvious promotion can be observed where AnomalyCD achieved average $F_1$ 30.27 compared to its bi-temporal version SAM-Bi 27.79, and ESIA with $F_1$ 41.58.

**Table. 2.** Comparison results about the anomaly localization maps from innate immune stage, evaluated on the global-scale GSA dataset.

| Dam break | | | | Burned area | | | | Fire point | | | | Input Type |
|---|---|---|---|---|---|---|---|---|---|---|---|---|
| Model | R | P | $F_1$ | Model | R | P | $F_1$ | Model | R | P | $F_1$ | |
| **Specialized anomaly localization models** | | | | | | | | | | | | |
| **MNDWI** | 74.64 | 37.33 | 40.37 | **dNDVI** | 39.01 | **87.32** | 48.24 | **TAI** | 2.91 | 14.81 | 3.92 | Reflectance |
| **MBWI** | 62.59 | 42.52 | 41.64 | **dNBR** | 43.30 | 82.05 | 52.50 | **NDFI** | 24.64 | 49.69 | 21.72 | Reflectance |
| **AWEI** | 69.21 | 43.73 | 43.44 | **RdNBR** | 63.97 | 67.10 | 62.13 | **AFD-$S^2$** | 2.33 | **89.95** | 4.51 | Reflectance |
| **General change detection models (bi-temporal)** | | | | | | | | | | | | |
| **CVA** | 33.65 | 58.90 | 29.82 | **CVA** | 15.90 | 71.42 | 23.56 | **CVA** | 40.68 | 36.97 | 31.35 | Reflectance |
| **SCG** | 62.83 | 43.04 | 38.97 | **SCG** | 38.36 | 62.01 | 42.11 | **SCG** | 60.21 | 33.05 | 31.87 | Reflectance |
| **MAD** | 48.48 | 56.05 | 35.77 | **MAD** | 26.78 | 71.66 | 35.73 | **MAD** | 52.16 | 34.30 | 33.22 | Reflectance |
| **SAM-Bi** | 65.94 | 46.43 | 43.76 | **SAM-Bi** | 50.04 | 56.23 | 47.18 | **SAM-Bi** | 48.13 | 22.58 | 22.62 | SAM Features |
| **General anomaly localization models (time-series)** | | | | | | | | | | | | |
| **COLD** | **91.06** | 39.27 | 44.71 | **COLD** | 58.45 | 64.39 | 47.01 | **COLD** | **67.62** | 23.69 | 22.79 | Reflectance |
| **EOAD** | 52.21 | 38.15 | 36.44 | **EOAD** | 19.98 | 52.21 | 24.54 | **EOAD** | 53.33 | 29.56 | 30.41 | Vegetation Index |
| **PKAD** | 71.15 | 39.58 | 28.69 | **PKAD** | **89.31** | 57.81 | **66.97** | **PKAD** | 62.68 | 17.35 | 20.27 | Vegetation Index |
| **AnomalyCD** | 56.70 | **66.20** | 52.24 | **AnomalyCD** | 46.92 | 68.16 | 49.38 | **AnomalyCD** | 38.37 | 35.79 | 29.63 | SAM Features |
| **ESIA** | 62.98 | 58.77 | **54.32** | **ESIA** | 82.36 | 56.04 | 63.31 | **ESIA** | 41.60 | 31.06 | **32.09** | HyperFree Features |

| Marine debris | | | | Oil leak | | | | Water bloom | | | | Average | | |
|---|---|---|---|---|---|---|---|---|---|---|---|---|---|---|
| Model | R | P | $F_1$ | Model | R | P | $F_1$ | Model | R | P | $F_1$ | R | P | $F_1$ |
| **Specialized anomaly localization models** | | | | | | | | | | | | | | |
| **FDI** | 62.08 | 20.93 | **19.44** | **NDOI** | 51.29 | 4.34 | 6.16 | **FAI** | 57.36 | 28.76 | 34.30 | - | - | - |
| **PI** | 22.68 | 5.89 | 6.38 | - | - | - | - | **SABI** | 61.21 | 39.62 | 36.93 | - | - | - |
| **General change detection models (bi-temporal)** | | | | | | | | | | | | | | |
| **CVA** | 46.35 | 7.81 | 6.56 | **CVA** | 8.61 | 4.65 | 6.03 | **CVA** | 12.15 | 45.09 | 19.09 | 26.22 | 37.47 | 19.40 |
| **SCG** | **84.77** | 0.73 | 1.43 | **SCG** | 43.05 | 6.45 | 10.52 | **SCG** | 52.15 | 29.72 | 29.91 | 56.90 | 29.17 | 25.80 |
| **MAD** | 52.43 | 4.82 | 5.13 | **MAD** | 10.16 | 4.41 | 6.10 | **MAD** | 19.23 | 42.07 | 26.30 | 34.87 | 35.55 | 23.71 |
| **SAM-Bi** | 18.09 | 10.42 | 11.60 | **SAM-Bi** | 53.37 | 11.45 | 18.47 | **SAM-Bi** | 41.25 | 18.20 | 23.10 | 46.14 | 27.55 | 27.79 |
| **General anomaly localization models (time-series)** | | | | | | | | | | | | | | |
| **COLD** | 70.34 | 4.55 | 4.48 | **COLD** | **69.64** | 3.90 | 7.21 | **COLD** | 51.79 | 33.61 | 33.91 | 68.15 | 28.24 | 26.69 |
| **EOAD** | 42.85 | 0.40 | 0.78 | **EOAD** | 32.62 | 10.38 | 12.91 | **EOAD** | 38.90 | 31.49 | 28.14 | 39.98 | 27.03 | 22.20 |
| **PKAD** | 67.29 | 0.99 | 1.83 | **PKAD** | 66.18 | 4.23 | 7.80 | **PKAD** | **83.95** | 19.80 | 28.57 | **73.43** | 23.29 | 25.69 |
| **AnomalyCD** | 28.07 | 10.09 | 14.36 | **AnomalyCD** | 44.01 | 7.63 | 12.87 | **AnomalyCD** | 55.50 | 16.49 | 22.72 | 44.93 | 34.06 | 30.27 |
| **ESIA** | 34.56 | **21.26** | 18.33 | **ESIA** | 38.73 | **18.11** | **20.03** | **ESIA** | 69.80 | **55.30** | **61.53** | 54.99 | **40.05** | **41.58** |

Putting the results in Table 2 together with the area statistic in Fig. 2(a), we find that anomalies with smaller areas are more difficult to localize such as the marine debris and oil leak with $F_1$ around 20.00, which are far behind the categories of dam break, burned area, and water bloom with $F_1$ around 60.00. This difference, however, does not consistent with the real performance since a few false pixels may reduce the quantitative results but have little performance on visualized maps as in Fig. 6 and Fig. 7. Taking the category of marine debris as example, the debris in Ground Truth occupies only a few pixels, and ESIA localizes most of them accurately. Due to the non-specific property in the innate immune stage, the unseen moving ship in the latest monitoring image is highlighted as well. Despite the tiny ship size, the boat would bring serve reduction in accuracy since it is several times the size of the debris.

We reported both the anomaly intensity maps and binary localization maps in Fig. 6 and Fig. 7 for six anomaly categories, where the intensity maps are continuous before the thresholding operation. Visualizing the continuous anomaly scores with gradient colors, it can be observed that the anomaly regions are clearly highlighted with a large response difference from surrounding objects, making the threshold selection easier. Please note that some localized objects may be also anomalous but computed as false alarms due to the contradiction of non-specific localized maps and anomaly-specific annotations. For example, in the bottom of localized map of water bloom category, the region in red box is located at the non-water body area and treated as false alarms in quantitative results, while it seems a kind of unobserved cropland changes needed to be noticed. ESIA has shown its generalization across different unseen regions and anomaly categories.

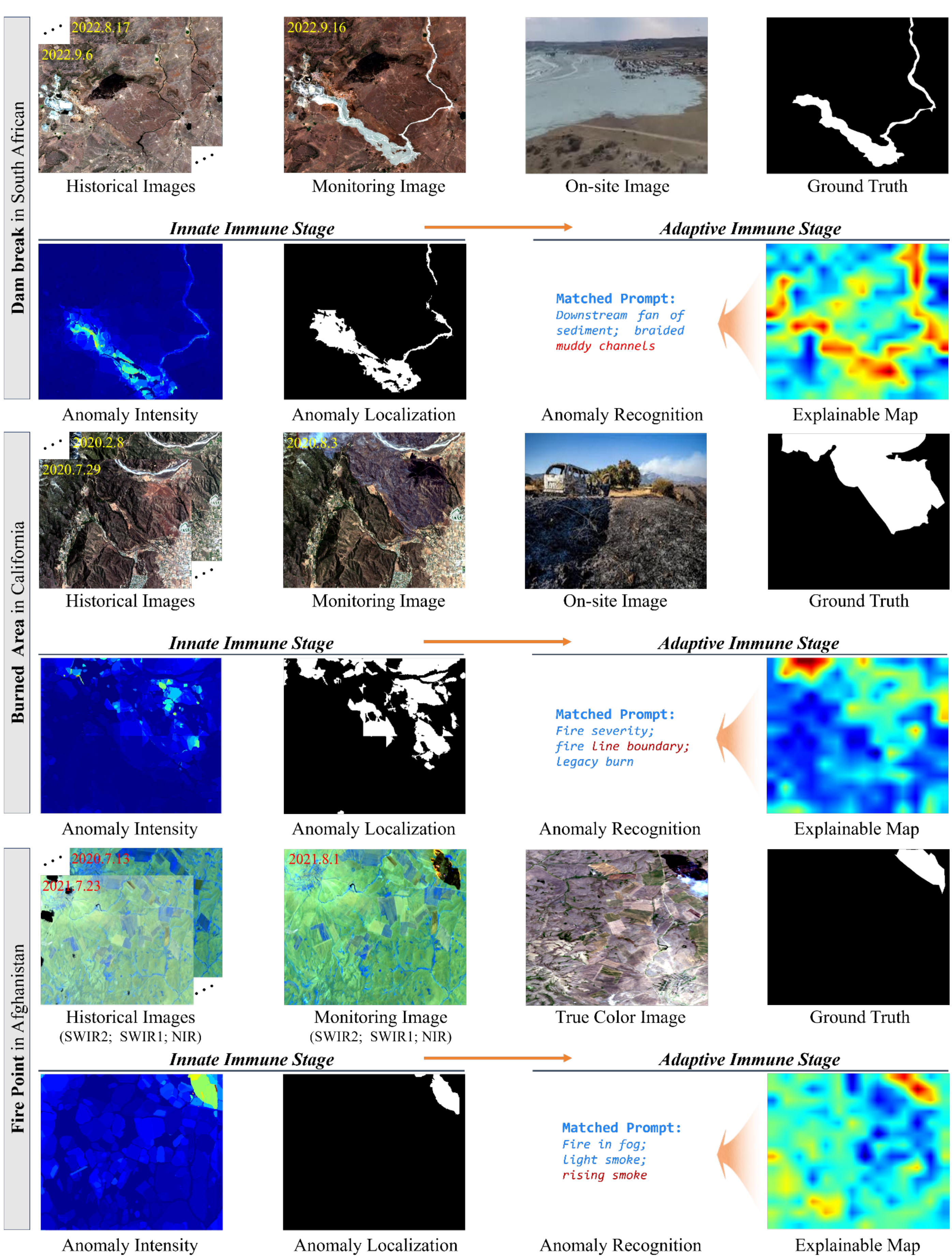

**Fig. 6.** Qualitative results of ESIA on the categories of dam break, burn area, and fire point, including both the innate and adaptive immune stages.

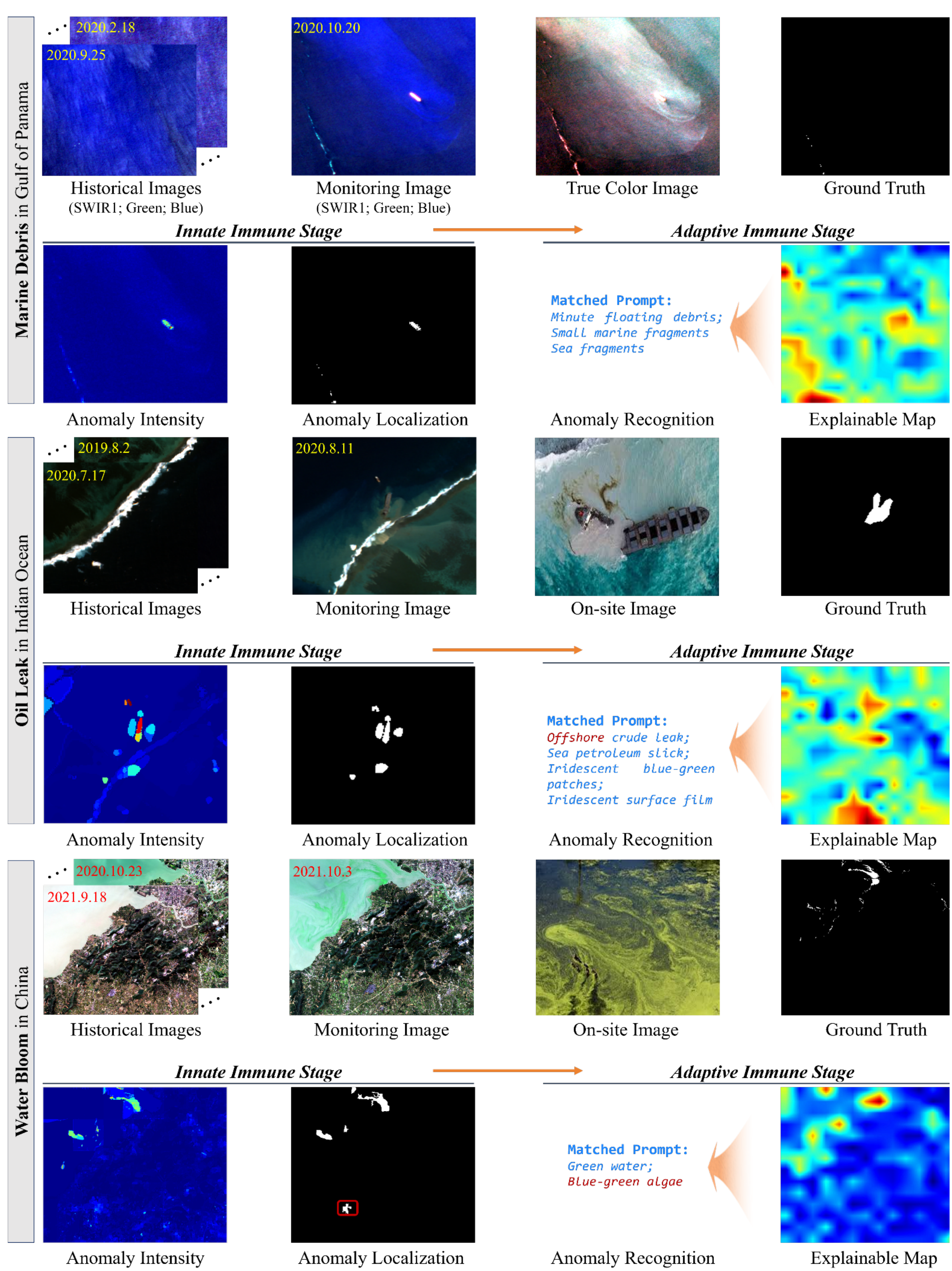

**Fig. 7.** Qualitative results of ESIA on the categories of marine debris, oil leak, and water bloom, including both the innate and adaptive immune stages.

*4.2.2 Benchmark Results at Adaptive Immune Stage*

Once innate immune localizes somewhere, adaptive immune then recognizes the corresponding anomaly category to be specific like BIS. This operation, however, is impossible for existing general models, especially referring to unseen categories and spectral images. Therefore, we only reported the comparison results with anomaly-specific models in Table 3, consistent with the models in innate immune stage. We convert them from localization to recognition by treating images with localized anomalies larger than some threshold as corresponding categories. Two indexes, FDI (Alì et al., 2024) and PI (Themistocleous et al., 2020) are not used for the category of marine debris since the extreme low-proportion makes the threshold infeasible. With appropriate conversion, these anomaly-specific indexes show $F_1$ score over 80.00 in many categories, and have achieved the optimal performance on Dam break with $F_1$ 85.71 by AWEI (Jiang et al., 2014), and on Fire point with $F_1$ 100.00 by NDFI (Liu et al., 2025). ESIA consistently shows the comparable or better results, with highest score on categories of burned area, oil leak, and water bloom. An interesting finding is that ESIA obtained precision 100.00 on many categories including dam break, fire point, oil leak, and water bloom, which implies that all the recognized samples were correct. The quantitative results prove the superiority of ESIA, even compared with anomaly-specific models.

**Table. 3.** Comparison results about the anomaly recognition from adaptive immune stage, evaluated on the global-scale GSA dataset.

| Dam break | | | | Burned area | | | | Fire point | | | |
|---|---|---|---|---|---|---|---|---|---|---|---|
| Model | R | P | $F_1$ | Model | R | P | $F_1$ | Model | R | P | $F_1$ |
| **Specialized anomaly recognition models** | | | | | | | | | | | |
| **MNDWI** | 100.00 | 60.00 | 75.00 | **dNDVI** | 88.24 | 68.18 | 76.92 | **TAI** | 44.44 | 66.67 | 53.33 |
| **MBWI** | 66.67 | 66.67 | 66.67 | **dNBR** | 70.59 | 80.00 | 75.00 | **NDFI** | 100.00 | 100.00 | 100.00 |
| **AWEI** | 75.00 | 100.00 | 85.71 | **RdNBR** | 82.35 | 60.87 | 70.00 | **AFD-S²** | 77.78 | 100.00 | 87.50 |
| **General anomaly recognition models** | | | | | | | | | | | |
| **ESIA** | 66.67 | 100.00 | 80.00 | **ESIA** | 88.24 | 78.95 | 83.33 | **ESIA** | 77.78 | 100.00 | 87.50 |
| **Marine debris** | | | | **Oil leak** | | | | **Water bloom** | | | |
| Model | R | P | $F_1$ | Model | R | P | $F_1$ | Model | R | P | $F_1$ |
| **Specialized anomaly recognition models** | | | | | | | | | | | |
| **-** | - | - | - | **NDOI** | 62.50 | 78.57 | 56.36 | **FAI** | 50.00 | 100.00 | 66.67 |
| | | | | - | - | - | - | **SABI** | 100.00 | 100.00 | 100.00 |
| **General anomaly recognition models** | | | | | | | | | | | |
| **ESIA** | 80.00 | 70.59 | 75.00 | **ESIA** | 66.67 | 100.00 | 80.00 | **ESIA** | 100.00 | 100.00 | 100.00 |

The adaptive immune of ESIA can not only recognize the anomaly category but also output explainable maps to tell the users why the category, and many open-vocabulary object attributes. Although these elements are difficult to be evaluated quantitatively, we show the related qualitative results in Fig. 6~7. We use the strategy proposed by Li et al. (2024) to generate the continuous explainable maps, where the class token is interacted with the image embeddings and the treats the similarity as the judge confidence. In Fig. 6 ~7, we can find there is a high consistence between the localized anomaly maps from innate immune and the explainable maps. Please note that localized anomaly maps have not been involved into the recognition process, and this consistency is a strong proof that ESIA understands the patterns and meanings of the categories in text-form. When some tiny inconsistency happens such as in the categories of marine debris and oil leak, users would be skeptical about the recognition results, and explainable maps act as some kind of confidence level from this point.

ESIA is compatible with open-vocabulary descriptions beyond the category such as the location, color, and the state. For each example in Fig. 6 ~7, we showed the related text descriptions with top few similarities with image embeddings, and we can find that the text-image matching mechanism actually knows more information than the expected. For example, the text “muddy channels” for dam break, “rising smoke” for fire point, “offshore” for oil leak, and “blue-green” for the water bloom. These descriptions are accurate enough and can provide addition information for users.

### *4.2.3 Comparison of Different Mutation Techniques*

ESIA imitates the mutation mechanism in BIS with the test-time tuning technique, which tunes a few embeddings to make the immune system adapt to properties of various categories efficiently. ESIA designs both changeable embeddings for the vision and text modalities and optimizes it with supervised, unsupervised and explainable losses. To show the design superiority, we compared it with some advanced test-time tuning techniques and reported the results in Fig. 8. All the techniques are based on MS-CLIP (Jakubik et al., 2024), the only multi-modality model for multispectral images, and we reported MS-CLIP as baselines. TPT (Shu et al., 2022) and CoOp (Zhou et al., 2022) only tune the text branch, while InCPL (Yin et al., 2024) lacks the step of negative selection and the explainable constrains. Without any tuning, MS-CLIP can only achieve the average $F_1$ 53.05, since the text descriptions about various anomalies are always rare in the pretrained corpus. CoOp, TPT, and InCPL achieved similar average performance around $F_1$ 70.00, while ESIA is the only model surpassed $F_1$ 80.00. ESIA has the largest lead in the category of oil leak, where the comparison techniques obtained $F_1$ 40.00 and ESIA with $F_1$ 80.00. InCPL has shown the best recognition ability on marine debris with $F_1$ 82.35 but failed to process the category of dam break, implying a certain instability. The improved accuracy, however, comes with the cost of increase time. Without the mutation process, the baseline speed of MS-CLIP is 0.7s/image, and the other three techniques have similar speed around 2.5s/image. In contrast, ESIA needs the highest time cost of 3.26s/image, where the image size is a fixed size of 224×224, equivalent to 5.02 $km^2$. ESIA increased the time cost and improved the accuracy.

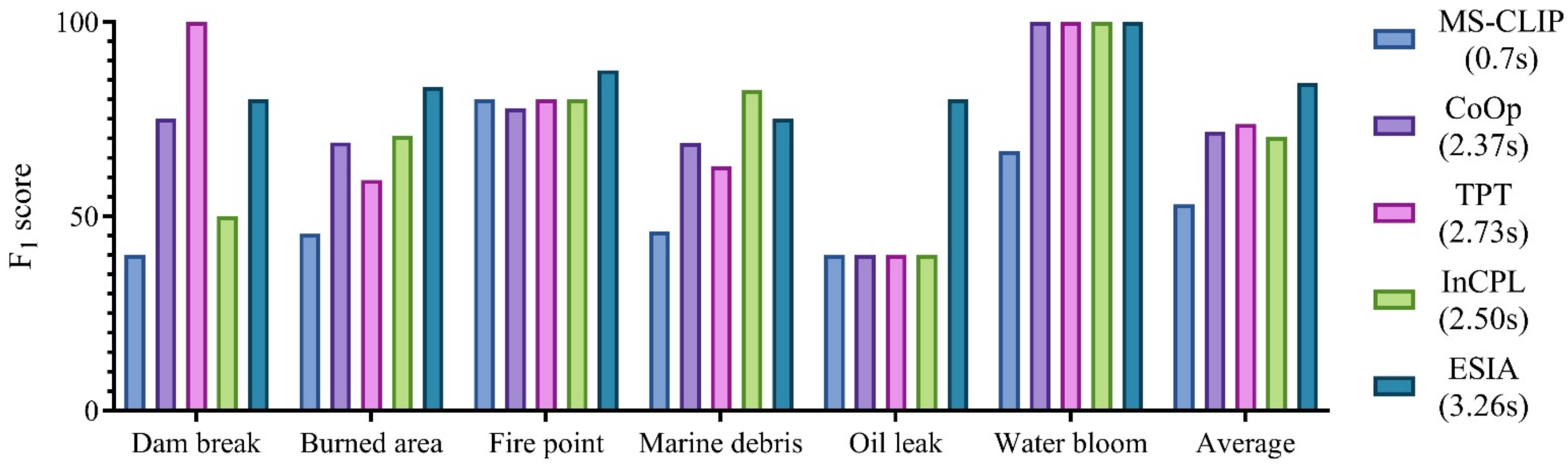

**Fig. 8.** Comparison of different test-time tuning techniques for the mutation in adaptive immune.

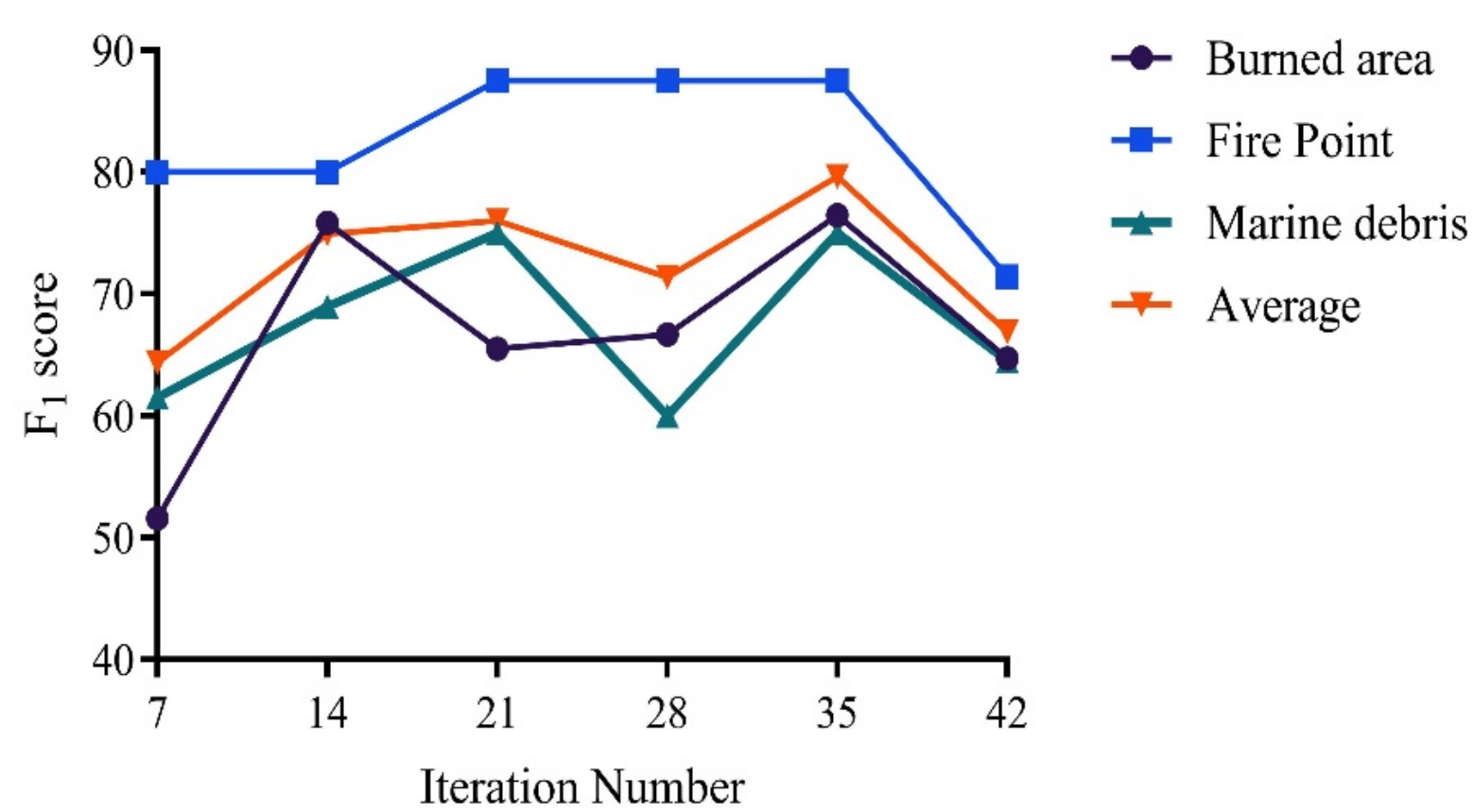

**Fig. 9.** Sensitivity of the iteration number on mutated embeddings.

**Table. 4.** Ablation of the key components in the adaptive immune stage.

| | Average | | |
|---|---|---|---|
| | R | P | F1 |
| w/o Negative selection | 78.89 | 77.09 | 77.54 |
| w/o $\beta_{image}$ | 78.89 | 81.58 | 79.05 |
| w/o $\beta_{text}$ | 66.67 | 81.25 | 72.26 |
| w/o $L_{bce}$ | 56.57 | 75.00 | 45.10 |
| w/o $L_{\exp}$ | 82.23 | 77.09 | 77.09 |
| w/o $L_{uns}$ | 75.56 | 80.56 | 77.09 |
| Oracle | 78.89 | 85.30 | 81.25 |

**Table. 5.** Sensitivity of the historical image number on the generated anomaly localization maps.

| | **Dam break** | | | **Burned area** | | | **Fire point** | | | **Oil leak** | | | **Water bloom** | | | **Average** | | |
|---|---|---|---|---|---|---|---|---|---|---|---|---|---|---|---|---|---|---|
| | R | P | $F_1$ | R | P | $F_1$ | R | P | $F_1$ | R | P | $F_1$ | R | P | $F_1$ | R | P | $F_1$ |
| **-4** | 59.18 | 46.68 | 46.97 | 73.76 | 60.79 | 62.48 | 44.48 | 29.55 | 31.37 | **38.96** | 14.81 | 17.66 | 62.05 | 53.21 | 57.11 | 55.69 | 41.01 | 43.12 |
| **-3** | 59.90 | 55.31 | 51.21 | 72.59 | 61.31 | 62.47 | **55.77** | 27.33 | 30.94 | 37.49 | 16.10 | 19.12 | 62.05 | 54.00 | 57.62 | 57.56 | 42.81 | 44.27 |
| **-2** | 59.17 | 57.42 | 51.49 | 71.82 | 63.11 | 62.95 | 55.54 | 30.76 | 34.26 | **38.96** | 18.89 | 20.44 | 58.11 | 60.66 | 58.25 | 56.72 | 46.17 | 45.51 |
| **-1** | 59.06 | 57.67 | 51.84 | 71.53 | **63.33** | 62.87 | 51.31 | **31.49** | **34.65** | **38.96** | **18.97** | **20.60** | 57.90 | **62.51** | 59.18 | 55.75 | **46.79** | 45.80 |
| **Base** | **62.98** | **58.77** | **54.32** | **82.36** | 56.04 | **63.31** | 41.60 | 31.06 | 32.09 | 38.73 | 18.11 | 20.03 | **69.80** | 55.30 | **61.53** | **59.07** | 43.81 | **46.22** |

*4.2.4 Sensitivity of Iteration Number in Mutation Process*

Iteration number is an important hyperparameter in the mutation process of ESIA, which decides the balance between maintaining the pre-trained general knowledge and adapting to the anomaly text phases for recognition. Intuitively, there should be a turning point as the iteration number increases. To find the point, we conducted sensitivity analysis the three most numerous anomalies including fire point and marine debris, and reported the results in Fig. 9. We grouped six iteration of supervised losses with one iteration of unsupervised loss, and thus conducted the statistics at intervals of 7. All the categories achieved stable promotion at the first 7 iterations but obvious fluctuations would come after that. The time to reach the peak varies with different types, where the category of burned area uses 14 iterations and the category of marine debris uses 21 iterations. We finally decided the hyperparameter setting according to the average performance, where the setting of 35 iterations reaches the highest $F_1$ 79.66.

*4.2.5 Ablation of Key Components in Adaptive Immune*

There are many key components that are designed to make ESIA understand the anomaly-related texts and unseen image properties, including the negative selection, changeable embeddings $\{\boldsymbol{\beta}_{text}, \boldsymbol{\beta}_{image}\}$, and specially designed loss terms. We treat the original ESIA results as the oracle version, and report the average ablation results in Table 4, using two more numerous categories of fire point and marine debris. The degrees of accuracy declines represent the effectiveness of corresponding components. The supervised loss term $L_{bce}$ shows the greatest importance, with $F_1$ 45.10 reduced from the oracle 81.25, although the remaining components keep existed. Without $L_{bce}$, the results either have recognized all the samples as anomalies or as normal ones, and cannot achieve an appropriate balance. By contrast, the elimination of other components seems showing similar performance with $F_1$ all above 70.00. Comparing the both mutable branches for text and image, $\boldsymbol{\beta}_{text}$ has shown more serious impacts than the $\boldsymbol{\beta}_{image}$, which has the difference around $F_1$ 6.79. We deduce that the mutable $\boldsymbol{\beta}_{text}$ may act as a role of correctness if negative selection is deleted, which can be proven by the $F_1$ 77.54 about the ablation of negative selection. Except for showing the degree of importance of each component, Table 4 exhibits that these designed strategies and loss terms can work together and benefit each other in adaptive immune stage.

*4.2.6 Effect of Time-series Observation in Innate Immune*

Innate Immune in ESIA uses time-series observation to locate the changed objects, and further filters out the disturb of normal changes (J. Li et al., 2024c). To quantitatively describe the effect of time-series images, we reduced the time steps successively from the farthest step from the latest image, and recomputed the corresponding anomaly maps. We test the reduction range from 1 to 4 considering the statistical data in Fig. 2(c), and the results are reported in Table 5. In the experiments of AnomalyCD (J. Li et al., 2024c) with high-resolution images from google Earth, the authors conclude that each drop in the number of historical normal time steps results in an average reduction of $F_1$ score metric by about 1.74 points evenly. In our experiments with sentinel 2 images, however, the reduction in $F_1$ caused by the decrease in steps does not seem to be uniform from the perspective of average performance. Specifically, by reducing 1- and 2-time steps, the $F_1$ score is reduced by 0.42 and 0.71 respectively, but if reducing 3 time phases, the $F_1$ score is directly reduced by 1.95. The $F_1$ score is reduced from 46.22 to 43.12 after deleting 4 time steps, where the impact is obviously growing. We deduce that the historical images of sentinel-2 satellite may have a certain degree of redundancy due to the low revisiting period, and reducing 1~2 time steps have little influence on the overall information. However, this robustness is limited and would be broken if more historical images are reduced, making ESIA lose the ability to remove the normal changes. The analysis experiments validate the promotion effect and unique trends of time-series observation on ESIA.

#### *4.2.7 Effect of Negative Selection on Text Prompts and Recognition Bias*

Inspired by BIS, we introduce the negative selection in ESIA to select appropriate spare prompts as the spare T/B cells in human body. To further demonstrate how the negative works in ESIA, we open the internal workflow and visualized the elements of $\{\mathbf{A},\mathbf{N}\}$, $\{Y_1,Y_2\}$, and $\{Y_1^s,Y_2^s\}$ in Fig. 10, taking examples from categories of burned area, marine debris, and fire point. The original $\{Y_1,Y_2\}$ contains hundreds of prompts generated by the Qwen3-Max model (Yang et al., 2025), and we show parts of it due to the space limit. Despite each anomaly category has its own prompts, we found all of them filtered anomaly prompts $Y_1$ that were professional such as the dNBR saturation, green-edge decline, and charred canopy, which are hard to understand for the general pre-trained multi-modality model MS-CLIP. The selected $Y_1^s$ belongs to the popular and common phrases including burned area, floating plastic, and forest fire. This difference in the degree of specialization can be visualized obviously in the demonstrated world cloud. For the prompts of normal classes, there is also a unified property that some phrases are frequently rejected by the three categories, such as no aberration, no defeat and anomaly-free, the forms of double negation. The remaining $Y_2^s$ consists of phrases such as regular cropland, normal farmland, and natural land. This property may be related to the weakness of CLIP-like models in capturing the inter-word relationship reported in the work of Long-CLIP (Zhang et al., 2024).

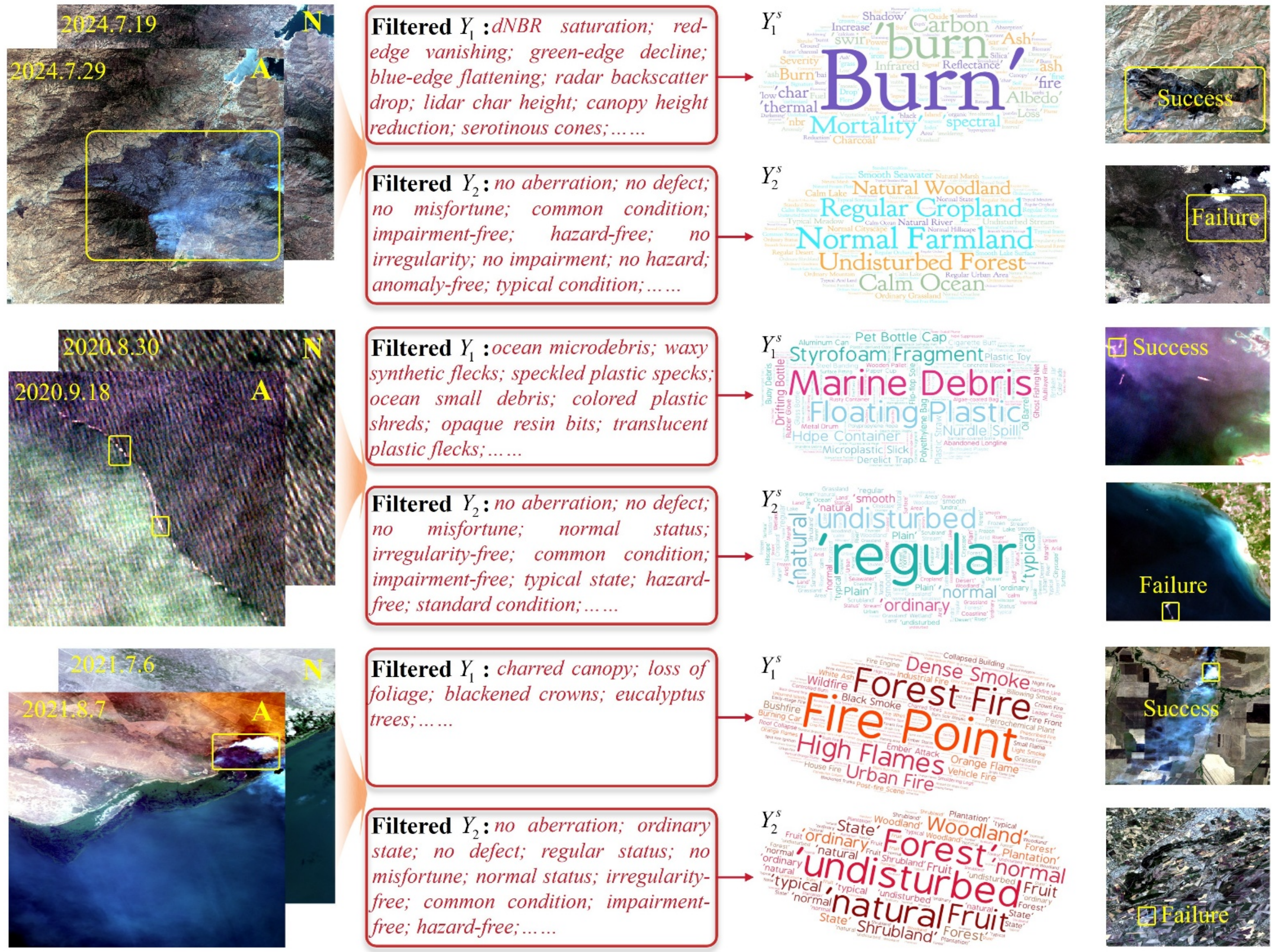

**Fig. 10.** Examples of the inner workflow in negative selection and corresponding successful and failure cases. We found professional anomaly prompts were more likely to be filtered out such as dNBR saturation, green-edge decline, and charred canopy. A high consistence can be clearly observed between the provided prompt images and the successful recognition probability.

Except for the effect on text prompts $\{Y_1, Y_2\}$, we also find the used $\{\mathbf{A}, \mathbf{N}\}$ decides the recognition bias of ESIA. Using the identical samples, we reported both successful and failure samples correspondingly in Figure X. The paired of $\{\mathbf{A},\mathbf{N}\}$ in the category of burned area has a slight smoke and a strong contrast between the burned area and the background. This property can be found similarly in the corresponding successful sample, while has a huge difference from the failure sample, with a large-scale black background. Similarly, the style that contains an area of purple and the style that has rising smoke seem the common property of successful samples for the marine debris and fire point, respectively. This high bias, decided by the given $\{\mathbf{A},\mathbf{N}\}$, is a double-edged sword, and needs to be processed carefully.

### *4.3 Application on Monitoring Degraded Agricultural Parcels in Dnipro Delta*

The collapse of Kakhovka Dam in June 2023, in the Russia-Ukraine war, led to widespread flooding, followed by severe soil salinization, erosion, and loss of arable land across the downstream delta region. The vital agricultural zone in southern Ukraine historically, faced a unprecedent challenge and large tracts of farmland would be unsuitable for cultivation. In this context, accurate assessment of degraded agricultural parcels is essential not only for understanding the environmental impact but also for guiding post-disaster recovery (Solomun et al., 2018). However, precise quantitative evaluation is rarely involved in related researches. This situation, motivates us to apply ESIA on the important application problem with large-scale satellite images, and give the answer about quantitative degraded parcels.

ESIA does not have seen the Dnipro delta or trained with related data, and it processes the problem directly with built innate immune and adaptive immune systems, which makes the task setting practical and also more challenging. We use the sliding window 1024 to infer the area covering 8704.46 km$^2$ at the innate immune stage, with images before and after the dam collapse of Kakhovka Dam. All the images are collected around July or August to ensure the identical cycle, and we have five images after the dam collapse to ensure the anomalies are permanent. We have shown the output anomaly localization map in Fig. 11, which is non-specific and contains various anomaly types beyond the degraded parcels. On the device GeForce RTX 4090, the innate immune stage only uses time 10.48 minutes, with speed 14.51 $km^2/s$. Since ESIA outputs object-level masks in innate immune stage, we can crop each object out with a certain range of surrounding context and feeds them into the adaptive immune for specific recognition tasks. After removing the objects with area less than 1000 pixels and taking 0.01 as threshold, 709 anomaly objects are processed to assign the specific attribute. Although ESIA is compatible with open-vocabulary text prompts, we focus on three types of degraded agricultural parcel, other agriculture anomalies, and the anomalies in the water in our context, where the text prompts are generated similarly as in GSA datasets. Under the test-time mutation setting, the adaptive immune takes 43.72 minutes to process all the localized objects. We have reported the used word clouds and recognized maps in Fig. 11. The statistical results show that there are 605 degraded agriculture parcels, covering 675.34 $km^2$, accounting for 14.43% of the agricultural area before the dam collapse. Due to the coverage problem of available images, the statistical results may be biased in spatial but can provide valuable reference for guiding post-disaster management.

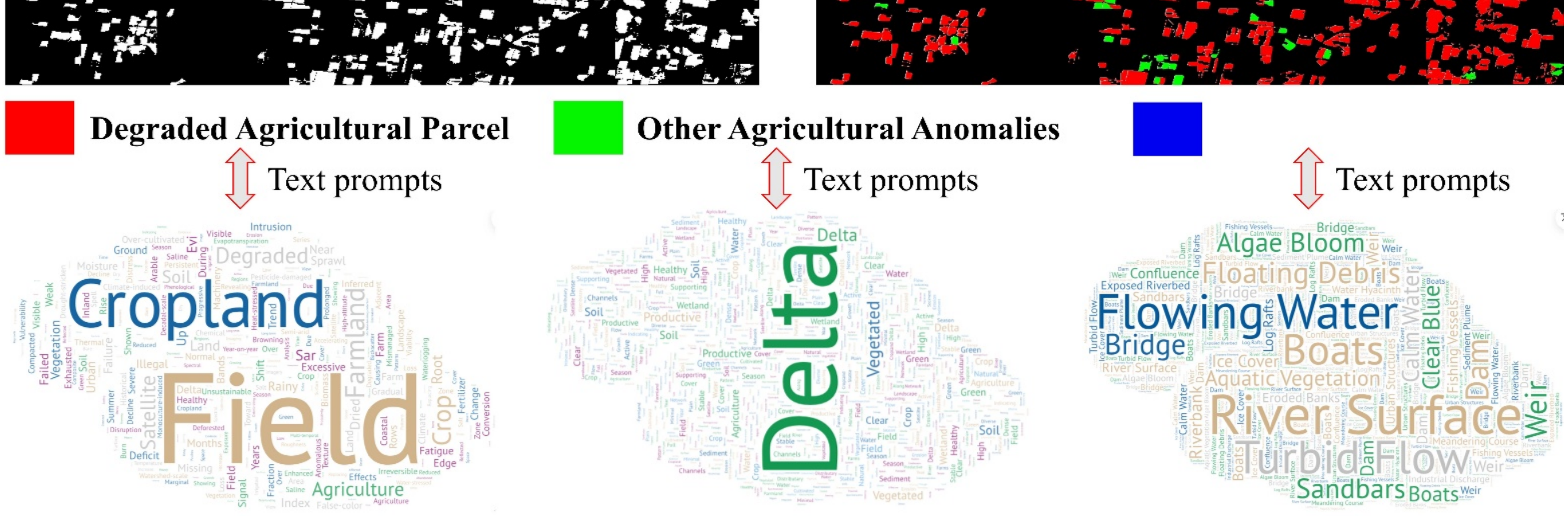

**Fig. 11.** As an unseen region of global concern, we apply ESIA to it to monitor the degraded agriculture parcels after the collapse of Kakhovka Dam in the Russia-Ukraine war. ESIA can accurately localize all the anomaly regions in innate immune and recognize the concerned degraded parcels in the adaptive immune in quantitative format.

### *4.4 Application on Monitoring Palisades Fire in Los Angeles, USA*

Palisades fire in Los Angeles is the source of tragic fire in Los Angeles in 2025. Accurate burned location and severity are needed for efficient fire resource management when the disaster occurs and the post-disaster recovery planning (Mohan, 2023). We apply ESIA to this real-world case to answer the questions of burned location and severity. The available sentinel-2 image, covering mainly the mountain scene, is largely different from the agricultural scene in Dnipro Delta, challenges the generalization of ESIA. At the innate immune stage, we use all the four pre-disaster images collected before January 2, 2025 and one post-disaster image collected at January 12, 2025, and infer the 237.13 $km^2$ with non-overlapped window size $1024 \times 1024$. Taking the identical threshold 0.01 with the case in Dnipro Delta, a binary localization map is generated in 75 seconds as in Fig. 12. Inputting the binary map into the adaptive immune stage, we use 10 text prompts for severe burn such as “high-severity fire damage” and 10 prompts for moderate burn such as “low-intensity fire impact”. The finally generated precise assessment map is also shown in Fig. 12. Considering the resolution of 10m/pixel, we can easily obtain the comprehensive assessment description as “Burned area: 79.8 km² (29.0 km² moderate, 50.8 km² severe)” for the policy-maker.

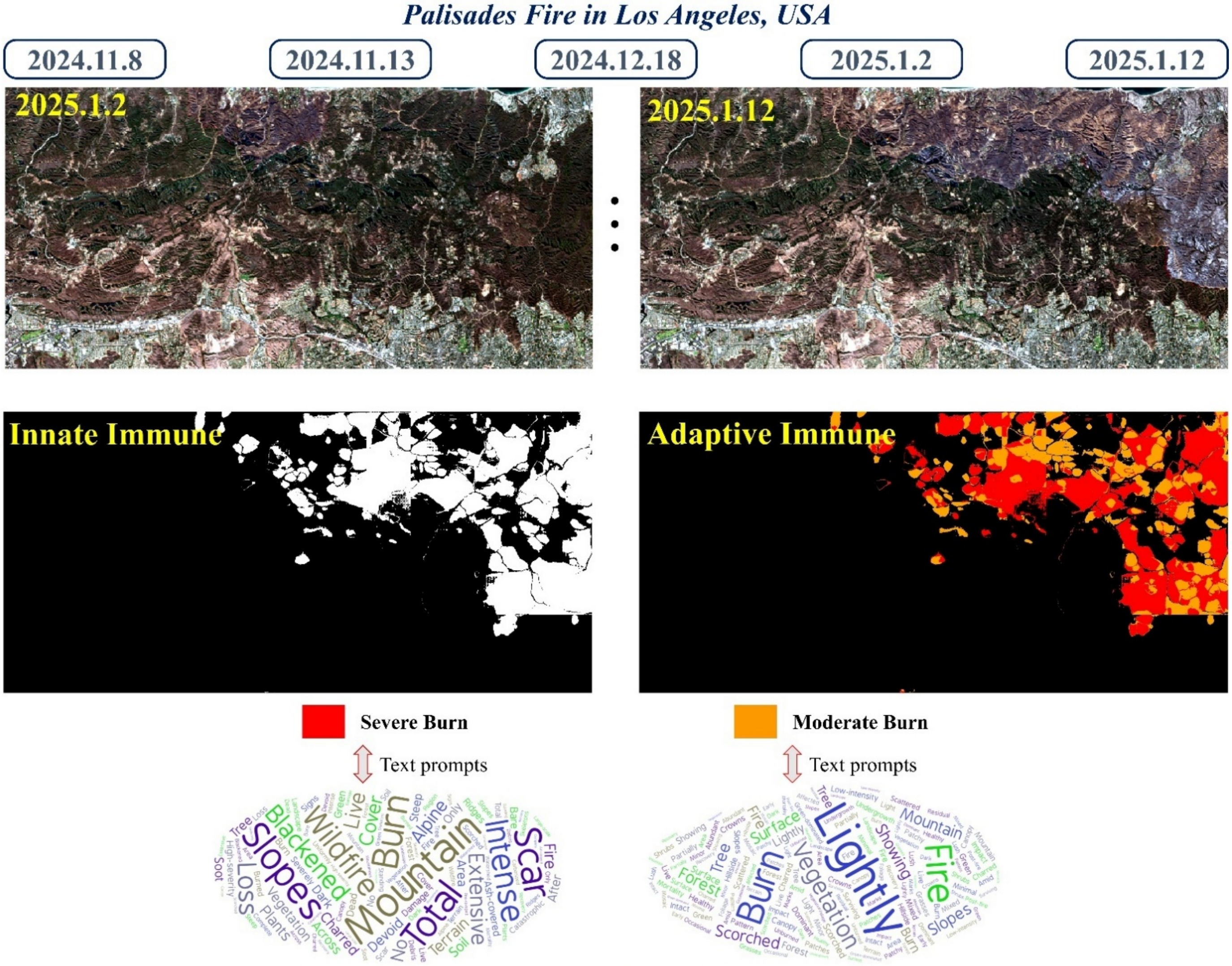

**Fig. 12.** We apply ESIA to monitor the burn severity in Palisades Fire in Los Angeles, USA. Similar to Dnipro Delta, ESIA has not seen the region before and shows generalize anomaly localization and fine-grained recognition ability.

## 5. Discussion

Proposed Earth surface immune system ESIA, firstly achieves a complete monitoring pipeline from localization to recognition. Applying ESIA to unseen global-scale GSA dataset covering 19801.60 km$^2$ and two regions of concern including Dnipro Delta and Palisades, it shows capability of generating accurate anomaly localization maps and open-vocabulary attribute recognition, as reported in Fig. 6~7 and Tables 2~3. This generalization ability, benefits from both the architecture philosophy from BIS (Liston et al., 2021; Shilts et al., 2022) and the many advanced techniques to support the implementation, including the spectral foundation model (Li et al., 2025), multi-modal pretraining (Jakubik et al., 2024), open-vocabulary recognition (Wu et al., 2024), and the test-time tuning (Shu et al., 2022). The evolution prior of BIS helps us build a complete anomaly monitoring mechanism to organize these techniques systematically.

ESIA has many properties consistent with the BIS to ensure the monitoring effectiveness. Except for the identical two-stage design, ESIA retains the properties of non-specificity and fast response in innate immune stage (Liston et al., 2021), and the properties of specificity and more accurate matching ability in adaptive immune (Shilts et al., 2022). Both stages are not totally broken but interactive, where the innate immune provides anomaly localization for adaptive immune to further check (Nguyen and Youn, 2025). Since changing anomalies always occupy a low proportion of Earth surface (J. Li et al., 2024c), this connection can avoid the adaptive immune to process all the regions and promote the overall efficiency largely. Additionally, ESIA retains the negative selection and mutation strategies as well in adaptive immune. These complex mechanisms, make ESIA not seem like a common end-to-end model in most studies (Sarkar et al., 2023; Zheng et al., 2024), but rather a slightly more complex system. Given the complexity of BIS after millions of years of evolution, this form of immune architecture may be the best choice under the tradeoff between simplicity and power.

Our implementation is not the only choice under the biological anomaly monitoring mechanisms. ESIA, as a framework, is expected to be prompted continuously with more advanced models. For example, the innate immune accepts the time-series observation images and outputs binary localization map, where the inner process can be instantiated with statistical models (Castillo-Villamor et al., 2021; Qiu et al., 2025a), traditional machine learning models, and foundation models (Li et al., 2025; Oquab et al., 2023). HyperFree (Li et al., 2025) is adapted to support the innate immune in our work, which shows amazing generalization but not the best choice for all categories such as the category of marine debris with extremely tiny sizes. Same thing with adaptive immune, it doesn't have to be instantiated with MS-CLIP (Jakubik et al., 2024) as the only way, some pre-trained models (Zheng et al., 2021), expert-rules (Clark et al., 2003), or more advanced multi-modal foundation model (Xu et al., 2025) are also allowed. The best implementation must be the one that meets the actual needs.

The shown generalization of ESIA is not the same as being able to do anything, and some limitations of ESIA is necessary to be discussed here.

(1) Difficulty in threshold selection. Without supervised samples, ESIA is free from certain anomaly categories (Chandola et al., 2009; Zheng et al., 2024). This gain, however, comes with the difficulty to select appropriate thresholds converting continuous anomaly degrees into a binary anomaly map. Prior work has shown the stability of fixed threshold for all the anomaly categories with high-resolution optical images (J. Li et al., 2024c), which is not observed in our study with medium-resolution and multi-spectral sentinel-2 images. Although some automatic thresholding methods such as Ostu are widely used (Jablonski and Mendrok, 2025; Seijo-Pardo et al., 2019), we found they were struggle to apply to large-scale images and can only output medium-level but not optimal results.

(2) Reliance on prompts samples. The designed mutation strategy in ESIA needs a single pair of samples, normal and anomaly patches, to tune the changeable parameters at test-time (Gross, 2025; Shu et al., 2022). Massive samples are impossible to collect while a pair of ones is easy to satisfy in practice. However, users should be aware of that large-sale pre-trained recognition models have ability to fast adapt to the given prompt samples (Ding et al., 2023), and the quality of given samples decides the bias of mutated parameters only with several iterations (Dong et al., 2024). Fig. 10 has proven this consistency, and it may cause failure recognition when anomaly pattern largely differs from the given prompt samples.

## 6. Conclusion

Motivated by the sophisticated art, biological immune system, we present an immune system ESIA for uncertain Earth surface anomalies from localization to attribute recognition. Extensive experiments have been conducted on constructed GSA dataset, consisting of time-series and global-scale multispectral images (19801.60 $km^2$), and tested ESIA in zero-shot manner without seeing the dataset before. Results show that ESIA has achieved better performance than a total of 22 comparing models that involves category-specialized models and general models with different image steps. Moreover, we apply ESIA directly to quantitatively assess the degraded agricultural parcels in Dnipro Delta, Russia and burned severity in Los Angeles, USA to solve application problems beyond the benchmark. Benefiting from the biological immune design and implementation with foundation models, ESIA extends the existing anomaly studies from localization to recognition, and from certain category to unfixed categories, which is more in line with the practical situation. Besides, ESIA processes unseen regions and images directly without a separate tuning process.

ESIA, acting as an Earth surface immune system, includes the stages of innate immune and adaptive immune, and key components of negative selection and mutation, following the spirit of BIS. This manuscript is not the only implementation of such system and more advanced foundation models or prior knowledge are promising to improve the performance. Moreover, how to design an automatic thresholding strategy for this zero-shot system is a challenging but important direction in the future, due to the great distribution difference for different scenes and anomaly categories. Despite the implementation difference, our ambition for the Earth surface anomalies is to locate it fast and recognize it accurately no matter where it happens.

**Acknowledgments**

This work was supported by the National Natural Science Foundation of China under Grant 424B2010 and Grant No. 42325105.